\pdfoutput=1
\documentclass[11pt]{article}
\usepackage[margin=1in]{geometry}
\usepackage[T1]{fontenc}
\usepackage{lmodern}
\usepackage{amsmath,amssymb,amsthm,mathtools}
\usepackage{graphicx,booktabs,microtype,array,needspace}
\usepackage{adjustbox,etoolbox}
\usepackage{placeins}
\makeatletter
\newcommand{\tableinplace}{}
\makeatother
\usepackage[backend=biber,style=alphabetic,maxalphanames=6,maxbibnames=99,natbib=true]{biblatex}
\DeclareFieldFormat[misc]{title}{\iffieldundef{eprint}{\mkbibemph{#1}}{\mkbibquote{#1\isdot}}}
\usepackage{algorithm,algorithmic}

\graphicspath{{./MS_FFT_plots/}}
\usepackage{xcolor}
\usepackage{tikz}
\usetikzlibrary{decorations.pathreplacing}
\usepackage[hidelinks]{hyperref}
\hypersetup{pdftitle={Fast Privacy Loss Accounting for Random Allocation via Multiscale FFT},pdfauthor={Vitaly Feldman and Moshe Shenfeld}}

\theoremstyle{plain}
\newtheorem{theorem}{Theorem}[section]

\newtheorem{lemma}[theorem]{Lemma}
\newtheorem{corollary}[theorem]{Corollary}
\newtheorem{claim}[theorem]{Claim}
\theoremstyle{definition}
\newtheorem{definition}[theorem]{Definition}
\theoremstyle{remark}
\newtheorem{remark}[theorem]{Remark}

\newcommand{\reals}{\mathbb R}
\newcommand{\naturals}{\mathbb N}
\newcommand{\eps}{\varepsilon}
\newcommand{\lvec}[1]{\reflectbox{$\vec{\reflectbox{$#1$}}$}}
\newcommand{\rem}[1]{\vec{#1}}
\newcommand{\add}[1]{\lvec{#1}}
\newcommand{\alg}{M}

\newcommand{\domain}{\mathcal X}

\newcommand{\dataset}{\boldsymbol s}
\newcommand{\view}{\boldsymbol v}

\newcommand{\lossRVfunc}[2]{L_{#1,#2}}
\newcommand{\dual}[1]{\mathcal D\left(#1\right)}
\newcommand{\alloc}[2]{\mathcal A_{#1}\left(#2\right)}
\newcommand{\privProf}[1]{\delta_{#1}}
\newcommand{\privProfRem}[1]{\rem{\delta}_{#1}}
\newcommand{\privProfAdd}[1]{\add{\delta}_{#1}}
\newcommand{\HockeyStick}[1]{\boldsymbol H_{#1}}
\newcommand{\HockeyStickFunc}[2]{\HockeyStick{#1}\left(#2\right)}
\newcommand{\HockeyStickDiv}[3]{\HockeyStick{#1}\left(#2\,\Vert\,#3\right)}
\newcommand{\expect}[2]{\underset{#1}{\mathbb E}\left[#2\right]}
\newcommand{\prob}[2]{\underset{#1}{\mathbb P}\left(#2\right)}
\newcommand{\domSt}{\preceq}
\newcommand{\upBnd}{U}
\newcommand{\lowBnd}{V}
\newcommand{\dfn}{\coloneqq}
\newcommand{\psiRem}[1]{\rem\psi_{#1}}
\newcommand{\psiAdd}[1]{\add\psi_{#1}}
\DeclareMathOperator{\supp}{supp}
\newcommand{\gridOf}[1]{\mathcal G_{#1}}
\newcommand{\grid}{\gridOf{m}}
\newcommand{\band}[1]{B_{#1}}
\newcommand{\spc}[1]{h_{#1}}
\newcommand{\proj}{\Pi}
\newcommand{\pref}[2]{\hat{#1}_{<#2}}
\newcommand{\distA}{\mu}
\newcommand{\distB}{\nu}
\newcommand{\msConv}{\textbf{ms-fft-merge}}
\newcommand{\selfConv}{\textbf{self-conv}}
\newcommand{\dyProj}{\textbf{dyadic-projection}}
\newcommand{\ctdAlloc}{\textbf{rand-alloc-ctd}}

\newcommand{\lr}{X}
\newcommand{\lrAvg}[1]{R_{#1}}
\newcommand{\lrHat}{\widehat R}
\newcommand{\sumHat}{\widehat S}

\newcommand{\ellLo}{\ell_-}
\newcommand{\ellHi}{\ell_+}
\newcommand{\indLo}{j_-}
\newcommand{\indHi}{j_+}
\newcommand{\lossPar}{\alpha'}
\newcommand{\wid}{\Delta}
\newcommand{\cx}{\le_{\mathrm{cx}}}
\newcommand{\ctd}{\mathrm{ctd}}

\newcommand{\ind}{\mathbf 1}
\newcommand{\defTag}[1]{\hypertarget{tag:#1}{(#1)}}
\newcommand{\refTag}[1]{\hyperlink{tag:#1}{(#1)}}
\newcommand{\eqTag}[2]{\overset{\text{\refTag{#1}}}{#2}}
\newcommand{\eqExp}[2]{\overset{(#1)}{#2}}

\title{Fast Privacy Loss Accounting for Random Allocation\\via Multiscale FFT}
\author{Vitaly Feldman \and Moshe Shenfeld\thanks{Work partially done while author was an intern at Apple.}}
\date{}

\begin{document}
\maketitle

\begin{abstract}
In the random allocation sampling scheme each user's data is used in $k$ steps chosen randomly and uniformly from a sequence of $t$ steps.
This scheme is known to improve on Poisson sampling \citep{CGHLKKMSZ24,CCGHST25} in private optimization and in communication-efficient aggregation \citep{AFKRT26}. The existing privacy loss distribution (PLD) accountant for this scheme runs in time quadratic in the inverse of the accuracy parameter $\alpha$ \citep{FS26pld}. We give an algorithm, \emph{multiscale FFT}, that represents likelihood ratios on a grid that is uniform within each dyadic interval and computes the exact rounding of every pairwise sum using FFTs within each interval. It keeps the guarantees of the accountant and improves the dependence of its running time on $\alpha$ from $O(\alpha^{-2})$ to $O(\alpha^{-1}\log(1/\alpha))$. In addition, we demonstrate that rounding can be performed via the mean-preserving connect-the-dots projection of \citet{DGKKM22}. This reduces the errors that rounding introduces in the mean of the privacy loss and in the $\chi^2$ divergence of the computed pair from the order of $\alpha$ to $O(\alpha^2/\log(t))$.
In our experiments, our algorithm achieves speedups of several orders of magnitude at high accuracy and gives improved privacy bounds for DP-SGD with PREAMBLE.
\end{abstract}

\section{Introduction}\label{sec:intro}
We consider privacy accounting for {\em random allocation}, a sampling scheme in which the data of each user is used in $k$ steps chosen randomly and uniformly from a sequence of $t$ steps (independently of the other users). This ensures that each user contributes a fixed number of times while still benefiting from the privacy amplification due to sampling. For $k=1$, this is the balls-and-bins sampling scheme used to define the batches in DP-SGD \citep{CGHLKKMSZ24,CCGHST25}. More generally, the scheme can be used to enforce a communication constraint in distributed aggregation by limiting the number of coordinate blocks that each user sends \citep{AFKRT26}. In both of these applications random allocation has been shown to have utility advantages over Poisson sampling \citep{CGHLKKMSZ24,CCGHST25,AFKRT26}. These applications have also motivated increasingly accurate analyses of its privacy guarantees \citep{FS25ra,DCO25,FS26pld}. For the Gaussian mechanism, the most accurate of these analyses shows that the privacy guarantees of random allocation are slightly better than those of Poisson sampling with the same noise scale and expected batch size in nearly all parameter regimes \citep{FS26pld}. Thus in these applications random allocation achieves a better privacy-utility trade-off than Poisson sampling.

As shown in \citep{FS26pld} (FS26 below), the privacy of random allocation can be analyzed using privacy loss distributions (PLDs). Given a PLD of a dominating pair of distributions of the per-step algorithm, their accountant computes a PLD that dominates random allocation applied to that algorithm (an {\em allocation PLD}). They reduce the allocation PLD to the distribution of a sum of $t$ independent likelihood ratios (that is, exponentiated privacy losses), and compute this distribution by repeated squaring. Each pairwise sum is computed by a direct convolution on a geometric grid and is rounded to the grid in the direction that preserves stochastic domination. We refer to this algorithm as {\em direct convolution}. Its running time is quadratic in the inverse of $\alpha$ {\textendash} the accuracy parameter in $\eps$ that bounds the error in the privacy loss. While direct convolution is efficient for a moderate level of accuracy, this dependence leads to a high running time in the high-accuracy regime. A highly accurate PLD is needed for the approximation error to remain small after a large number of compositions. For example, the analysis of DP-SGD with PREAMBLE in FS26 includes thousands of inner compositions and more than ten thousand training updates. In this regime, computing a PLD for which the final $\eps$ is accurate to within $0.05$ typically takes more than an hour on a single Apple M4 core. Finding the noise scale that achieves a target $\eps$ requires multiple such computations, and a careful choice of the accuracy parameters of direct convolution.

The standard FFT-based approach to the composition of PLDs \citep{KJH20,GLW21} does not resolve this problem. Composition adds privacy losses, whereas random allocation adds their exponentials, that is, likelihood ratios. A uniform grid over likelihood ratios needs to resolve small ratios over a range that can be very large. As a result, ensuring accuracy $\alpha$ using FFT requires an impractically large grid, and smaller grids have been demonstrated to have a poor trade-off between accuracy and running time \citep[Appendix D]{FS26pld}. Thus, given the practical benefits of random allocation, we address the question of whether there is a more efficient approach to privacy accounting for random allocation.

\subsection{Our Contribution}\label{sec:contribution}
Our main contribution is a {\em multiscale FFT} algorithm with {\em connect-the-dots projections}. This algorithm computes a dominating allocation PLD in time that is nearly linear in the inverse of the accuracy parameter.  Our algorithm approximates the PLD of the same dominating pair of distributions as in FS26, accepts a general PLD as its input and outputs a PLD. In particular, the output PLD supports additional accounting operations such as subsampling and composition.

\paragraph{Multiscale FFT.} We replace the geometric grid of FS26 by a grid with $m$ uniformly spaced points in each dyadic interval $[2^j,2^{j+1})$ for $j \in \mathbb{Z}$, and compute every pairwise sum by multiscale FFT (Section~\ref{sec:multiscale}). Its output is identical to rounding the exact sum onto the grid.
For the same accuracy, the dyadic grid has a comparable number of points. Its main advantage is for computation of each merge: direct convolution's runtime is quadratic in grid size since it sums all pairs of grid points, whereas multiscale FFT takes time nearly linear in the number of grid points. As in FS26, the input PLD is only approximated between its $\beta/2t$ and $1-\beta/2t$ quantiles, where $\beta$ bounds the probability of the tails (translating to an increase in $\delta$). Let $\wid$ be the length of the smallest interval that contains these quantiles of the log-likelihood ratio $\ln(P(\omega)/Q(\omega))$ (where $(P,Q)$ is the dominating pair that defines the input PLD) both for $\omega\sim P$ and for $\omega\sim Q$, that is, of the input PLD and of the negation of its dual. Then for $t$ steps the running time of our algorithm (up to reading the input; Corollary~\ref{cor:ctd-original} and Remark~\ref{rem:continuous}) is
\begin{equation}\label{eq:intro-runtime}
 O\!\left(\frac{(\wid+\log(t))\log^2(t)}{\alpha}\log\!\left(\frac{\log(t)}{\alpha}\right)\right).
\end{equation}
For example, for the Gaussian mechanism with noise scale $\sigma$, $\wid=O(1/\sigma^2+\sqrt{\log(t/\beta)}/\sigma)$. Thus, for a fixed number of steps, the running time is linear in $\wid/\alpha$ up to a logarithmic factor, whereas the running time of direct convolution is quadratic in the analogous ratio \citep[Theorem~4.6]{FS26pld}.

\paragraph{Connect-the-dots projections.} Conservative rounding in FS26 moves every value in the same direction, and thus increases the expected privacy loss and the $\chi^2$ divergence of the computed pair by amounts of the order of $\alpha$. We refer to the increase in the expected privacy loss as the {\em bias}.
We replace it by the connect-the-dots (CtD) projection of \citet{DGKKM22}, which splits the mass of each value between its two neighboring grid points so as to preserve its mean. We demonstrate that the bias of our algorithm and its error measured as the increase in the $\chi^2$ divergence of the computed pair both drop to $O(\alpha^2/\log(t))$, with the same running time and the same $(\alpha,\beta)$ guarantee (Section~\ref{sec:ctd}). Our focus on measuring the increase in $\chi^2$ relies on recent work showing that  $\chi^2$ of the dominating pair tightly controls the privacy loss for (subsampled) composition and for random allocation \citep{FS26clt}. Our algorithm applies CtD projection at multiple levels and, as we argue briefly in Remark~\ref{rem:lower-levels}, the results in \citep{FS26clt} apply to all but a small number of the highest-level projections.

CtD projection gives valid upper bounds only when it is applied to likelihood ratios under $Q$. Therefore we compute only the add direction and obtain the remove direction as its dual.

\paragraph{Numerical results.} Numerical evaluation using an implementation that can be found at \citep{PLD_acc_msfft} shows that our algorithm is faster than direct convolution of FS26 by several orders of magnitude in the regimes that require high accuracy. As a quick numerical example, in the setting of DP-SGD with PREAMBLE above, multiscale FFT computes the allocation PLD in $0.012$ seconds and, for the same noise scale, improves the upper bound on $\eps$ at $\delta=10^{-6}$ from $0.985$ to $0.92$ (Section~\ref{sec:preamble}).
Naturally, the speedups are smaller when the input PLD is concentrated, since direct convolution is already fast on the resulting small grid. We report the full range of these comparisons in Section~\ref{sec:experiments}, together with improved privacy bounds for DP-SGD with PREAMBLE after Poisson subsampling and composition.

We also examine the effect of the CtD projection on the accuracy of the final bound and compare it with the error resulting from the rounding approach in FS26. Our evaluation suggests that the CtD improves the dependence of the running time on the accuracy from about $1/\alpha$ to about $\sqrt{1/\alpha}$ (Section~\ref{sec:experiments}). This suggests that the accuracy of the final bound behaves essentially like the bias and $\chi^2$ error.

\subsection{Technical Overview}\label{sec:overview}
We represent distributions of likelihood ratios on a grid that is uniform within each dyadic interval $[2^j,2^{j+1})$ (or {\em band}) and whose spacing doubles between consecutive bands (Figure~\ref{fig:grid}). Such a grid gives a relative approximation at every scale while allowing the use of FFT-based convolutions within a band. The main step of our algorithm computes the distribution of the sum of two independent random variables supported on this grid and projects it back onto the same grid using the CtD projection (we refer to this step as a {\em merge}).

Computing a merge by direct convolution requires going over all pairs of grid points, whereas FFT requires both values to lie on the same uniform grid. The obvious issue with using FFT on the dyadic grid is that two values from different bands lie on grids with different spacings.

However, the sum of such values does not require the finer resolution of the lower band. Specifically, for a value $x$ in band $j$ and a value $y$ in a lower band, the sum $x+y$ lies in bands $j$ or $j+1$, and the spacing of the grid above $2^j$ is at least the spacing of band $j$. We therefore project $y$ onto the uniform grid with the spacing of band $j$ (extended down to zero), on which $x$ also lies. A crucial question is whether this can be done without increasing the approximation error. We show that the answer is yes as long as the uniform grids of different bands are nested, which holds for the dyadic grid (every grid point in band $j$ or above is a multiple of the spacing of band $j$). In fact, in this case the result is identical to the projection of $x+y$ itself onto the grid.

Thus for every band $j$ we apply this projection to all the mass of each distribution below band $j$. The resulting {\em projected prefix} occupies only $m+1$ grid points, regardless of the number of lower bands. All the pairs of values in which the larger value is in band $j$ then lie on the same uniform grid. Their contribution to the sum is given by three convolutions on this grid (band $j$ of both distributions with each other, and band $j$ of each distribution with the projected prefix of the other), which we compute using FFTs of length $4m$. Pairs in which both values lie below band $j$ are accounted for, at a finer resolution, in a lower band. Overall, every pair of values is accounted for exactly once. Therefore a merge computes the projection of the exact sum in time that is nearly linear in the size of the grid (Theorem~\ref{thm:merge}).

\begin{figure}[t]
\centering
\begin{tikzpicture}[x=0.8cm,y=0.72cm,
  knot/.style={circle,fill,inner sep=0pt,minimum size=3.4pt},
  latt/.style={circle,draw,fill=white,inner sep=0pt,minimum size=4.2pt},
  pref/.style={circle,draw,fill=black!40,inner sep=0pt,minimum size=4.2pt},
  lab/.style={font=\scriptsize},
  side/.style={font=\scriptsize,anchor=east}]
\draw[->] (-0.3,1.7) -- (16.7,1.7);
\foreach \x in {0,0.5,0.625,0.75,0.875,1,1.25,1.5,1.75,2,2.5,3,3.5,4,5,6,7,8,10,12,14,16}
  \node[knot] at (\x,1.7) {};
\foreach \x in {0,1,2,4,8,16} \node[lab,below=2pt] at (\x,1.7) {$\x$};
\foreach \a/\b/\j in {0.5/1/-1,1/2/0,2/4/1,4/8/2,8/16/3}{
  \draw[decorate,decoration={brace,amplitude=2.5pt}] (\a,1.88) -- (\b,1.88);
  \node[lab] at ({(\a+\b)/2},2.3) {$\j$};}
\node[side] at (-0.45,2.3) {band};
\node[side] at (-0.45,1.7) {grid $\gridOf{4}$};
\draw[->,thin,shorten <=2pt,shorten >=3pt] (2.5,1.7) -- (2,0);
\draw[->,thin,shorten <=2pt,shorten >=3pt] (2.5,1.7) -- (3,0);
\node[lab] at (2.95,1.0) {$y$};
\foreach \x in {4,5,6,7,8,10,12,14} \draw[densely dashed,gray] (\x,1.12) -- (\x,0.14);
\draw[->] (-0.3,0) -- (16.7,0);
\foreach \x in {0,...,15} \node[latt] at (\x,0) {};
\foreach \x in {0,...,4} \node[pref] at (\x,0) {};
\node[side] at (-0.45,0) {grid $h_2\mathbb Z_{\ge0}$};
\draw[decorate,decoration={brace,mirror,amplitude=3pt,raise=5pt}] (0,0) -- (4,0)
  node[lab,midway,below=8pt,align=center] {mass below band $2$\\ (projected prefixes)};
\draw[decorate,decoration={brace,mirror,amplitude=3pt,raise=5pt}] (4.1,0) -- (14,0)
  node[lab,midway,below=8pt,align=center] {support of the three convolutions at band $2$};
\end{tikzpicture}
\caption{The dyadic grid and the merge step for $m=4$. Top: the grid $\gridOf{4}$ and its bands $\band j=[2^j,2^{j+1})$ (bands $-2,-3,\ldots$ accumulate at zero), where band $2$ is $[4,8)$ with spacing $\spc 2=1$. Bottom: the uniform grid of band $2$ extended down to zero. A value $y$ below band $2$ is projected onto this grid (arrows), after which the mass below band $2$ occupies $m+1$ points. The knots of $\gridOf{4}$ in $[4,16)$ (dashed, in bands $2$ and $3$) are a subset of this grid, and $[4,16)$ contains the support of the three convolutions at band $2$ (band $2$ of both inputs with each other, and band $2$ of each input with the projected prefix of the other). Thus projecting the sums onto $\gridOf{4}$ gives the same measure as projecting the exact sums (Theorem~\ref{thm:merge}).}
\label{fig:grid}
\end{figure}

As in FS26, we compute the distribution of the sum of $t$ identically distributed likelihood ratios by repeated squaring, which requires only logarithmically many merges. Each projection changes a value by a factor of at most $1+1/m$, resulting in an error of at most $1/m$ in the log space. Adding nonnegative values preserves a common bound on their multiplicative error. Thus the error of the computed sum grows with the number of merges that a summand passes through (at most $\lceil\log_2(t)\rceil$) and not with the number of summands. In particular, $m=O(\log(t)/\alpha)$ points per band suffice for accuracy $\alpha$. Every distribution in the computation has $O(\wid+\log(t))$ bands (a merge adds at most one band at the top), and each band is merged in $O(m\log(m))$ time, which gives the running time in~\eqref{eq:intro-runtime}.

Viewed as a random map, the CtD projection preserves the conditional mean of the likelihood ratio, $\expect{}{\widetilde Y\mid Y}=Y$, where $\widetilde Y$ is the projection of a random variable $Y$. This makes the increase of the $\chi^2$ divergence caused by a projection quadratic in the spacing of the grid, due to the cancellation of the first-order errors (Section~\ref{sec:chi2}), as for the mean and the variance of the privacy loss \citep[Section~6]{FS26clt}. With a spacing of the order of $\alpha/\log(t)$ and $O(\log(t))$ levels of projections, this gives the bound $O(\alpha^2/\log(t))$ on the $\chi^2$ error and on the bias stated above. This is in contrast to the rounding of FS26, which moves every value in the same direction and shifts the expected privacy loss by $O(\alpha)$.

At the same time, the CtD projection moves some of the mass down and need not give stochastic domination. Instead, preservation of the conditional mean increases $Y$ in the convex order, which is preserved under sums of independent random variables. Under $Q$, both privacy profiles of random allocation are expectations of convex functions of the average likelihood ratio, which gives the validity of both directions (Lemma~\ref{lem:validity}).

As discussed, we obtain the remove direction as the dual of the add direction. For tightness we show that in both directions the exact and the computed losses can be coupled so that their finite values differ by at most $\gamma=(\lceil\log_2(t)\rceil+1)\ln(1+1/m)$. To construct the coupling, we implement each projection as a random map that moves a value to one of its two neighboring grid points with the probabilities that preserve its mean (Section~\ref{sec:tightness}). For an input with an unbounded range of likelihood ratios (such as the Gaussian mechanism) we first truncate its tails, at a negligible cost in the tail parameter, and then project it onto the grid by one more CtD projection (Remark~\ref{rem:continuous}).

\subsection{Related Work}\label{sec:related}
The dominating pair of distributions for random allocation with $k=1$ and Gaussian noise was given in the analyses of balls-and-bins sampling for DP-SGD \citep{CGHLKKMSZ24} and for the matrix mechanism \citep{CCGHST25}. It was extended in \citep{FS25ra} to algorithms whose steps are dominated by an arbitrary pair of distributions, and \citet{DCO25} derived it independently for Gaussian noise. \citet{CGHLKKMSZ24} and \citet{CCGHST25} estimate the privacy parameters of this pair using Monte Carlo simulations, and \citet{DG26} base Monte Carlo accounting for $b$-min-sep subsampling (a generalization of balls-and-bins sampling) in banded matrix factorization via dynamic programming. These simulations provide strong evidence that the privacy guarantees of $1$-out-of-$t$ allocation for Gaussian noise are similar to those of $t$ compositions of Poisson sampling with rate $1/t$. The primary issue with Monte Carlo estimation is its computational cost when $\delta$ is small and a high confidence level is required. In addition, it estimates the privacy parameters of the pair rather than its full privacy profile or PLD, and thus does not support subsequent accounting operations such as subsampling and composition.

The analysis in \citep{FS25ra} gives provable bounds on the privacy parameters of $k$-out-of-$t$ allocation via a reduction to a $k$-fold composition of $1$-out-of-$t/k$ allocation (Lemma~\ref{lem:reduction}), including bounds on the R\'enyi divergence of the dominating pair. \citet{DCO25} and \citet{SK26} give bounds on the R\'enyi divergence for Gaussian noise and for matrix mechanisms, respectively, the latter computed by dynamic programming. The primary disadvantage of bounds on R\'enyi divergence is that their conversion to $(\eps,\delta)$ bounds is known to be somewhat lossy. Additionally their bounds apply only to R\'enyi orders of at least $2$ and therefore provide a loose bound on the expected privacy loss.

Our work builds on the PLD accountant of FS26, which computes the PLD of the dominating pair numerically and thus supports subsequent accounting operations. Specifically, we use its representation of the allocation PLD via sums of likelihood ratios, its repeated squaring schedule (which gives logarithmic dependence on the number of steps), and its accounting operations on PLDs. We improve the cost of each merge at the resolution needed for an accurate PLD, reduce the $\chi^2$ error and the bias by replacing the rounding with the CtD projection, and obtain the remove direction as the dual of the add direction. In the remove direction, FS26 also merge a likelihood ratio under $P$ and thus discretize both the input PLD and its dual. Since stochastic domination is not preserved by the dual, they discretize the dual separately and compute it from the exact input \citep[Section~4 and Remark~C.2]{FS26pld}.

Numerical composition of PLDs was introduced by \citet{MM18} and \citet{SMM19}, who quantize the PLD into {\em privacy buckets} and compose them by direct convolution, using exponentiation by squaring for a logarithmic dependence of the running time on the number of compositions. \citet{KJH20,KJPH21,KH21} compose discretized PLDs using FFT, and \citet{GLW21} give a tight analysis of the dependence of the running time on the accuracy. This approach is now the standard for the analysis of DP-SGD and is supported by several libraries \citep{Google20,Microsoft21,Meta21}. \citet{GKKM22} give a faster self-composition algorithm in which the grid spacing and the truncation interval grow with the number of composed copies, so that each of its stages uses only polylogarithmically many grid points. These algorithms compute the distribution of a sum of privacy losses. In contrast, random allocation requires adding likelihood ratios, and the dyadic grid is designed to address their large dynamic range.

The mean-preserving projection we use is the pessimistic CtD approximation of \citet{DGKKM22}, expressed in terms of likelihood ratios (Claim~\ref{clm:ctd-pessimistic}). Viewed as a random map, it is the one-dimensional case of the dual quantization of \citet{PW12}, which preserves the conditional mean and therefore gives upper bounds on the expectations of convex functions and preserves the convex order. \citet{PW09} apply dual quantization recursively to the partial sums of a random walk and bound the resulting error for smooth functions.

The closest technique to our multiscale FFT merge is the fast projected convolution of \citet{Hac07}, which computes the exact $L^2$-orthogonal projection of the convolution of two piecewise-polynomial functions onto a locally refined grid (for example, one refined towards zero) in nearly linear time, under certain conditions on the grids. As in our algorithm, the grids are decomposed into levels of uniform grids, and the projection is incorporated before the discrete convolutions on each level, which are computed by FFT. Our merge applies this principle to the CtD projection of atomic measures, for which the nesting of the uniform grids of the bands makes the projection of the lower bands exact (Theorem~\ref{thm:merge}).

\section{Preliminaries}\label{sec:prelims}
We follow the notation of FS26. The statements that we use, with the proof of the new identity~\eqref{eq:alloc-dual}, appear in Appendix~\ref{apd:prelims}. For a random variable $Y$ we denote its distribution by $f_Y$ and its complementary CDF by $\overline F_Y(x)\dfn\prob{}{Y>x}$. Random variables take values in $[-\infty,\infty]$. We denote by $\ln$ and $\log_2$ the natural and binary logarithms, and write $\log$ inside $O(\cdot)$, where the base is immaterial.

\paragraph{Random allocation and adjacency.} We consider {\em $t$-step algorithms}, which apply an algorithm $\alg$ to $t$ subsets of the dataset in sequence, where each step also receives the outputs of the previous steps (its {\em view}). In the {\em $k$-out-of-$t$ random allocation} scheme, each element of the dataset is added to $k$ of the $t$ subsets, chosen uniformly at random and independently of the other elements. We denote the resulting algorithm by $\alloc{t,k}{\alg}$, and by $\alloc t\alg$ when $k=1$. To define adjacency for sampling schemes that operate on a fixed number of elements, the domain $\domain$ of the elements is augmented with a {\em null element} $\bot$, which $\alg$ treats as absent, that is, $\alg((\dataset,\bot),\view)=\alg(\dataset,\view)$ for every dataset $\dataset$ and view $\view$. Two datasets are {\em adjacent} if one of them is obtained from the other by replacing a single element by $\bot$. Thus replacing an element by $\bot$ has the same effect on the output as removing it, while the number of elements, and thus the sampling scheme, remain unchanged.

\paragraph{Privacy loss distributions and privacy profiles.} For distributions $P$ and $Q$ over a domain $\Omega$, the {\em privacy loss random variable} $\lossRVfunc PQ$ is $\ln(P(\omega)/Q(\omega))$ for $\omega\sim P$, and its distribution is the {\em privacy loss distribution (PLD)} \citep{DR16}. For $\kappa\ge0$, the {\em $\kappa$-hockey-stick divergence} between $P$ and $Q$ is $\HockeyStickDiv{\kappa}{P}{Q}\dfn\expect{}{(1-\kappa e^{-\lossRVfunc PQ})_+}$, where $(x)_+\dfn\max\{0,x\}$ \citep{BKOZB12}. For a random variable $L$ we refer to $\HockeyStickFunc{\kappa}{L}\dfn\expect{}{(1-\kappa e^{-L})_+}$ as its {\em hockey-stick functional}. The {\em privacy profile} $\privProf{\alg}(\eps)$ of an algorithm $\alg$ is the largest $e^\eps$-hockey-stick divergence between its output distributions on two adjacent datasets with the same view \citep{BBG18}. Since the hockey-stick divergence is asymmetric, we denote by $\privProfRem{\alg}$ the privacy profile in the {\em remove} direction, in which the second dataset is obtained from the first by replacing an element by $\bot$, and by $\privProfAdd{\alg}$ the one in the {\em add} direction, in which the first dataset is obtained from the second in this way. The algorithm is {\em $(\eps,\delta)$-differentially private} if $\privProf{\alg}(\eps)=\max\{\privProfRem{\alg}(\eps),\privProfAdd{\alg}(\eps)\}\le\delta$ \citep{DKMMN06}.

\paragraph{Dominating pairs.} We say that a PLD $\lossRVfunc{P_1}{Q_1}$ {\em dominates} a PLD $\lossRVfunc{P_2}{Q_2}$ if $\HockeyStickDiv{\kappa}{P_2}{Q_2}\le\HockeyStickDiv{\kappa}{P_1}{Q_1}$ for all $\kappa\ge0$. A pair $(P,Q)$ is a {\em dominating pair} for $\alg$ in the remove (add) direction if $\privProfRem{\alg}(\eps)\le\HockeyStickDiv{e^\eps}{P}{Q}$ (respectively, $\privProfAdd{\alg}(\eps)\le\HockeyStickDiv{e^\eps}{P}{Q}$) for all $\eps$ \citep{ZDW22}. Domination is transitive, and it is preserved under composition, whose PLD is the sum of the PLDs of the components \citep[Theorem~10]{ZDW22}. If every step of a $t$-step algorithm $\alg$ is dominated by $(P,Q)$ in the remove direction, then $\alloc t\alg$ is dominated by $(\bar P_t,Q^t)$ in the remove direction and by $(Q^t,\bar P_t)$ in the add direction, where
\[
 \bar P_t\dfn\frac1t\sum_{i=1}^tQ^{i-1}\times P\times Q^{t-i},
\]
and the $k$-out-of-$t$ allocation reduces to compositions of such allocations (Lemma~\ref{lem:reduction}). We refer to the PLDs $\lossRVfunc{\bar P_t}{Q^t}$ and $\lossRVfunc{Q^t}{\bar P_t}$ as the {\em allocation PLDs}.

\paragraph{PLD realizations.} A random variable $L$ is a {\em PLD realization} if $\expect{}{e^{-L}}\le1$ and $\prob{}{L=-\infty}=0$ \citep[Definition~3.1]{FS26pld}. Every PLD is a PLD realization, and conversely, every PLD realization is the PLD of some pair of distributions. The {\em dual} $\dual L$ of a PLD realization $L$ is the PLD of the reversed pair, so that $\dual{\lossRVfunc PQ}=\lossRVfunc QP$ (\ref{clm:pld-real}). Its distribution is that of $-L$ with probabilities weighted by $e^{-L}$, with mass $1-\expect{}{e^{-L}}$ at $+\infty$ \citep[Definition~3.1]{FS26pld}. A PLD realization $\upBnd$ {\em dominates} a PLD realization $L$ if $\HockeyStickFunc{\kappa}{L}\le\HockeyStickFunc{\kappa}{\upBnd}$ for all $\kappa\ge0$, and in this case $\dual\upBnd$ dominates $\dual L$ (\ref{clm:dual-dom}). The privacy profiles of $L$ and of $\dual L$ are both expectations of functions of $e^{-L}$:
\begin{equation}\label{eq:profiles}
 \HockeyStickFunc{e^\eps}{L}=\expect{}{(1-e^{\eps-L})_+},
 \qquad
 \HockeyStickFunc{e^\eps}{\dual L}=1-\expect{}{e^{-L}}+\expect{}{(e^{-L}-e^\eps)_+}.
\end{equation}
The first formula is the definition, and the second one holds since the mass of $\dual L$ at $+\infty$ contributes $1-\expect{}{e^{-L}}$ and its finite part contributes $\expect{}{e^{-L}(1-e^\eps e^{L})_+}=\expect{}{(e^{-L}-e^\eps)_+}$. Both functions are convex in $e^{-L}$. For the PLD $L=\lossRVfunc QP$ in the add direction, $e^{-L}=P(\omega)/Q(\omega)$ for $\omega\sim Q$ is the likelihood ratio under $Q$.

\paragraph{The allocation PLDs.} For a PLD realization $L$ let
\begin{equation}\label{eq:psi}
 \psiRem t(L)\dfn\ln\biggl(\frac1t\Bigl(e^{L^{(0)}}+\sum_{i=1}^{t-1}e^{-\dual L^{(i)}}\Bigr)\biggr),
 \qquad
 \psiAdd t(L)\dfn-\ln\Bigl(\frac1t\sum_{i=1}^te^{-L^{(i)}}\Bigr),
\end{equation}
where the superscripts denote independent copies. FS26 show that $\lossRVfunc{\bar P_t}{Q^t}=\psiRem t(\lossRVfunc PQ)$ and $\lossRVfunc{Q^t}{\bar P_t}=\psiAdd t(\lossRVfunc QP)$, and that these transformations preserve domination (Theorem~\ref{thm:psi} and \citep[Section~4]{FS26pld}). In $\psiRem t$, the factor $e^{L^{(0)}}$ is the likelihood ratio under $P$ and the other factors are likelihood ratios under $Q$. Thus FS26 compute the remove direction from both the input PLD and its dual. However, the two allocation PLDs are the PLDs of the same pair in the two directions. Therefore, for every PLD realization $L$ (\ref{clm:alloc-dual}),
\begin{equation}\label{eq:alloc-dual}
 \psiRem t(\dual L)=\dual{\psiAdd t(L)}.
\end{equation}
Thus both allocation PLDs are determined by the average $\lrAvg t\dfn\frac1t\sum_{i=1}^te^{-L^{(i)}}$ of $t$ likelihood ratios under $Q$, since $\psiAdd t(L)=-\ln(\lrAvg t)$. By~\eqref{eq:profiles}, both privacy profiles of random allocation are expectations of convex functions of $\lrAvg t$.

\paragraph{Stochastic domination and tightness.} For any two random variables $\lowBnd, \upBnd$, we write $\lowBnd\domSt\upBnd$ if $\overline F_\lowBnd\le\overline F_\upBnd$, and $\lowBnd\domSt_{(\alpha,\beta)}\upBnd$ if $\overline F_\lowBnd(x)\le\overline F_\upBnd(x-\alpha)+\beta$ for all $x$. Following FS26 we say that $\upBnd$ {\em $(\alpha,\beta)$-tightly stochastically dominates} $\lowBnd$ if $\lowBnd\domSt\upBnd\domSt_{(\alpha,\beta)}\lowBnd$. Stochastic domination implies domination, and $\lowBnd\domSt_{(\alpha,\beta)}\upBnd$ implies $\HockeyStickFunc{e^\eps}{\lowBnd}\le\HockeyStickFunc{e^{\eps-\alpha}}{\upBnd}+\beta$ (\ref{clm:stoch-hs}). The output of our accountant dominates the exact PLD but need not stochastically dominate it. Therefore we state its guarantees directly in terms of privacy profiles.

\begin{definition}[Tight domination]\label{def:tight-dom}
Given $\alpha\ge0$ and $\beta\in[0,1]$, a PLD realization $\upBnd$ {\em $(\alpha,\beta)$-tightly dominates} a PLD realization $L$ if for every $\eps\in\reals$,
\[
 \HockeyStickFunc{e^\eps}{L}\le\HockeyStickFunc{e^\eps}{\upBnd}\le\HockeyStickFunc{e^{\eps-\alpha}}{L}+\beta.
\]
\end{definition}
That is, $\upBnd$ dominates $L$, and the privacy profile of $\upBnd$ is at most that of $L$ shifted by $\alpha$, plus $\beta$. By the implication above, if $\upBnd$ dominates $L$ and $\upBnd\domSt_{(\alpha,\beta)}L$ then $\upBnd$ $(\alpha,\beta)$-tightly dominates $L$, and in particular tight stochastic domination implies tight domination with the same parameters (\ref{clm:stoch-hs}).

\begin{definition}[$\chi^2$ error and bias]\label{def:chi2}
For a PLD realization $L$ we denote $\chi^2(L)\dfn\expect{}{e^{L}}-1\in[0,\infty]$. For PLD realizations $\upBnd$ and $L$ with $\chi^2(L)<\infty$, the {\em $\chi^2$ error} of $\upBnd$ relative to $L$ is $\chi^2(\upBnd)-\chi^2(L)$. The {\em bias} of $\upBnd$ relative to $L$ is the increase $\expect{}{\upBnd\ind_{\{\upBnd<\infty\}}}-\expect{}{L\ind_{\{L<\infty\}}}$ in the expectation of the finite part of the loss.
\end{definition}
Note that $\chi^2(L)\ge0$, since $\expect{}{e^{L}}\ge1/\expect{}{e^{-L}}\ge1$ by Jensen's inequality, and that $\chi^2(L)=\infty$ if $L$ has mass at $+\infty$. For a pair $(P,Q)$ with $P\ll Q$, $\chi^2(\lossRVfunc PQ)=\expect{\omega\sim Q}{(P(\omega)/Q(\omega))^2}-1$ is the $\chi^2$ divergence of $P$ from $Q$. The $\chi^2$ divergence of a single step controls the Gaussian approximation of the privacy losses of random allocation \citep[Section~5]{FS26clt}.

\section{Multiscale FFT}\label{sec:multiscale}
In this section we describe our accountant (Algorithm~\ref{alg:ctd-alloc}) and the multiscale FFT merge that it relies on. The accountant computes the add direction from the distribution of the sum of $t$ independent likelihood ratios under $Q$, and it obtains the remove direction as the dual of the result (by~\eqref{eq:alloc-dual}). After projecting the likelihood ratio of its input onto a grid that is uniform within each dyadic interval, it then computes the distribution of the sum by repeated squaring, in which every convolution is followed by the projection onto the grid (a merge). We define the grid and the projection in Section~\ref{sec:ctd-projection} and describe the accountant in Section~\ref{sec:ctd-alloc}. In Section~\ref{sec:ms-merge} we show that the multiscale FFT merge computes every merge exactly in time nearly linear in the size of the grid (Theorem~\ref{thm:merge}). We analyze the accountant in Section~\ref{sec:ctd}.

\subsection{The Grid and the Projection}\label{sec:ctd-projection}
For an integer $m\ge2$, let
\[
    \grid\dfn\{0\}\cup\biggl\{2^j\left(1+\frac {i-1}{m}\right):j\in\mathbb Z,\, i\in[m]\biggr\}.
\]
We refer to its points as {\em knots}, to $\band j\dfn[2^j,2^{j+1})$ as {\em band} $j$, to $\spc j\dfn2^j/m$ as the spacing of band $j$, and to $\spc j\mathbb Z_{\ge0}$ as the {\em lattice} of band $j$. The knots in band $j$ are $m$ consecutive points of its lattice (Figure~\ref{fig:grid}). The analysis uses two properties of the grid (Claim~\ref{clm:grid}):
\begin{itemize}
\item[\defTag{G1}] \emph{Resolution}: Every $x>0$ satisfies $u\le x\le v$ for two consecutive knots $0<u<v\le(1+1/m)u$.
\item[\defTag{G2}] \emph{Dyadic scaling}: For every $j\in\mathbb Z$, every knot in band $j$ or above is an integer multiple of $\spc j$.
\end{itemize}
The multiscale FFT merge relies on the scaling, and the analysis of the accuracy in Section~\ref{sec:ctd} on the resolution.

Our algorithm relies on CtD projections \citep{DGKKM22} onto $\grid$, which split the mass of each value between its two neighboring knots so as to preserve its mean. Formally, for a set of knots $S\subseteq[0,\infty)$ that contains $0$ and a value $x$ between adjacent knots $u<v$ of $S$, let
\[
 \proj_S(\delta_x)\dfn\frac{v-x}{v-u}\,\delta_u+\frac{x-u}{v-u}\,\delta_v,
\]
and let $\proj_S(\delta_x)\dfn\delta_x$ for $x\in S\cup\{\infty\}$. We extend $\proj_S$ linearly to finite measures and abbreviate $\proj\dfn\proj_{\grid}$ and $\proj_h\dfn\proj_{h\mathbb Z_{\ge0}}$ for $h>0$ (Definition~\ref{def:projections}).

\paragraph{Properties of the projections.} Our analysis uses only the following properties of the projections. We state them, and Theorem~\ref{thm:merge}, for general projections, since it is possible to use the multiscale FFT merge with rounding to get lower bounds (Remark~\ref{rem:merge-rounding}). A {\em projection} onto a set of knots $S$ is a map from finite measures on $[0,\infty]$ to finite measures on $S\cup\{\infty\}$ that fixes every measure supported on $S\cup\{\infty\}$. Let $\xi$ and $\zeta$ be finite measures on $[0,\infty]$, $h>0$ and $j\in\mathbb Z$.
\begin{itemize}
\item[\defTag{P1}] \emph{Locality}: $\proj_S(\delta_x)$ is supported on $\{u,v\}$ for every $x$ between adjacent knots $u<v$ of $S$.
\item[\defTag{P2}] \emph{Linearity}: $\proj_S(\xi+\zeta)=\proj_S(\xi)+\proj_S(\zeta)$.
\item[\defTag{P3}] \emph{Commutativity}: $\proj_h(\xi*\zeta)=\proj_h(\xi)*\zeta$ if $\zeta$ is supported on $h\mathbb Z_{\ge0}$.
\item[\defTag{P4}] \emph{Projection hierarchy}: $\proj_{2h}(\proj_h(\xi))=\proj_{2h}(\xi)$, and $\proj(\proj_{\spc j}(\xi))=\proj(\xi)$ if $\xi$ is supported on $[2^j,\infty)$.
\item[\defTag{P5}] \emph{Mean preservation}: $\proj_S(\delta_x)$ has mean $x$ for every $x\ge0$.
\end{itemize}
The multiscale FFT merge relies on \refTag{P1}--\refTag{P4}, and the analysis of the accountant in Section~\ref{sec:ctd} on \refTag{P1} and \refTag{P5}. The CtD projections satisfy all five properties (Lemma~\ref{lem:lattice}). The second part of the projection hierarchy is specific to the dyadic grid and relies on its scaling and the fact $2^j\in\grid$.

\subsection{The Accountant}\label{sec:ctd-alloc}
Algorithm~\ref{alg:ctd-alloc} receives a PLD realization $L$ in the add direction. (For a PLD realization $\rem L$ in the remove direction we use $L=\dual{\rem L}$, which only reweights its atoms.) It chooses the number $m$ of points per band from the accuracy parameter (line~1; see Corollary~\ref{cor:ctd-original}) and projects $e^{-L}$ onto $\grid$, which gives a PLD realization $L'$ for which $e^{-L'}$ is supported on $\grid$ (lines~2--3). It then computes by repeated squaring the distribution $f_{\sumHat}$ of the computed sum $\sumHat$, which approximates the sum of $t$ independent copies of $e^{-L}$ (line~4). Finally, it returns $\add L_t\dfn-\ln(\sumHat/t)$ and its dual $\rem L_t\dfn\dual{\add L_t}$ (line~5). That is, $\add L_t$ has mass $f_{\sumHat}(z)$ at $-\ln(z/t)$, and $\rem L_t$ has mass $\frac zt f_{\sumHat}(z)$ at $\ln(z/t)$ for every $z>0$ and the remaining mass $1-\expect{}{\sumHat/t}$ at $+\infty$.

\begin{algorithm}[ht]
\caption{CtD accountant \ctdAlloc$(L,t,\alpha)$}\label{alg:ctd-alloc}
\begin{algorithmic}[1]
\REQUIRE $L$ a PLD realization (add direction) with bounded finite values; $t\in\naturals$, $t\ge2$; $\alpha\in(0,1]$
\ENSURE $\add L_t$ and $\rem L_t$ that $(\alpha,0)$-tightly dominate $\psiAdd t(L)$ and $\psiRem t(\dual L)$
\STATE $m\gets\lceil(\lceil\log_2(t)\rceil+1)/\alpha\rceil$
\STATE $\indLo\gets\lfloor\ellLo/\ln(2)\rfloor$; $\indHi\gets\lfloor\ellHi/\ln(2)\rfloor$
\STATE $L'\gets\dyProj(L;\indLo,\indHi,m)$ \COMMENT{$\supp(e^{-L'})\subseteq\grid$}
\STATE $f_{\sumHat}\gets\selfConv(f_{e^{-L'}},t)$
\STATE $\add L_t\gets-\ln(\sumHat/t)$; $\rem L_t\gets\dual{\add L_t}$
\RETURN $\add L_t$, $\rem L_t$
\end{algorithmic}
\end{algorithm}

\paragraph{Discretizing the input.}\label{sec:input} In line~2 of Algorithm~\ref{alg:ctd-alloc}, $\ellLo$ and $\ellHi$ are the infimum and the supremum of the finite values of $-L$ (an input with unbounded losses is first truncated, as described in Remark~\ref{rem:continuous}). Since band $j$ contains $e^x$ for $j=\lfloor x/\ln(2)\rfloor$, the indices $\indLo$ and $\indHi$ are the bands that contain $e^{\ellLo}$ and $e^{\ellHi}$, the infimum and the supremum of the positive values of the likelihood ratio $e^{-L}$, and $\indHi-\indLo\le(\ellHi-\ellLo)/\ln(2)+1$. The range $[2^{\indLo},2^{\indHi+1}]$, from the start of band $\indLo$ to the end of band $\indHi$, contains every positive value of $e^{-L}$, and the discretization \dyProj{}$(L;\indLo,\indHi,m)$ returns the PLD realization $L'$ for which $f_{e^{-L'}}=\proj(f_{e^{-L}})$, that is, the CtD projection of the likelihood ratio onto $\grid$. It applies to discrete and continuous inputs alike and needs only the probability and the first moment of $e^{-L}$ on each cell between adjacent knots.

\paragraph{Repeated squaring.} As in FS26, we compute the $n$-fold self-convolution of a distribution $\distA$ by repeated squaring, and replace every convolution by a merge, which we compute by the multiscale FFT merge \msConv{} (Algorithm~\ref{alg:ms-conv}). Let $n=\sum_{k=1}^s2^{b_k}$ with $b_1<\cdots<b_s$, let $\distA^{(0)}\dfn\distA$ and $\distA^{(b)}\dfn\msConv(\distA^{(b-1)},\distA^{(b-1)})$ for $1\le b\le b_s$, and let $\xi_1\dfn\distA^{(b_1)}$ and $\xi_k\dfn\msConv(\xi_{k-1},\distA^{(b_k)})$ for $2\le k\le s$. We denote the output $\xi_s$ by $\selfConv(\distA,n)$. Repeated squaring uses $b_s+s-1\le2\lceil\log_2(n)\rceil$ merges. Since a convolution is the distribution of a sum of independent random variables, the computation corresponds to a binary tree with $n$ leaves, which we refer to as the {\em unfolded tree}. Its leaves are independent random variables with distribution $\distA$, and every internal node is a merge applied to the values computed at its two children (a merge of $\distA^{(b-1)}$ with itself is a node whose children are two independent copies of the subtree of $\distA^{(b-1)}$). The depth of this tree is $\lceil\log_2(n)\rceil$ (Claim~\ref{clm:depth}).
Unlike FS26, we do not truncate the tails between merges, since a merge adds at most one band (Theorem~\ref{thm:merge}).

\subsection{The Multiscale FFT Merge}\label{sec:ms-merge}
We now describe the multiscale FFT merge (Algorithm~\ref{alg:ms-conv}) and show that it is {\em exact}, that is, that it computes the projection onto $\grid$ of the exact convolution of its inputs. It accounts for every pair of values in the band of the larger value, and it replaces the mass of each input below that band by its projection onto the lattice of the band. This projection does not change the result (Theorem~\ref{thm:merge}), and it reduces the contribution of each band to three convolutions on its lattice, which we compute by FFTs.

For a probability distribution $\distA$ on $\grid\cup\{\infty\}$, let $\distA_j$ be its restriction to $\band j$, $\distA_{<j}$ its restriction to $(0,2^j)$, $\distA_+$ its positive finite part, and $\distA(0)$ and $\distA(\infty)$ its masses at $0$ and at $\infty$. We refer to $\pref\distA j\dfn\proj_{\spc j}(\distA_{<j})$ as its {\em projected prefix}. By locality \refTag{P1}, $\pref\distA j$ is supported on $\{0,\spc j,\ldots,m\spc j\}$, regardless of the number of bands below $j$, since it moves every value in $(0,2^j)=(0,m\spc j)$ to the neighboring knots of $\spc j\mathbb Z_{\ge0}$. Algorithm~\ref{alg:ms-conv} computes the projection of the convolution $\distA*\distB$ of two such distributions band by band, and it maintains the projected prefixes of the current band in $\hat\distA$ and $\hat\distB$ (line~5).

\begin{algorithm}[ht]
\caption{Multiscale FFT merge \msConv$(\distA,\distB)$}\label{alg:ms-conv}
\begin{algorithmic}[1]
\REQUIRE $\distA,\distB$ probability distributions on $\grid\cup\{\infty\}$, with $\distA_+,\distB_+$ supported on $\bigcup_{j=j_{\min}}^{j_{\max}}\band j$
\ENSURE $\proj(\distA*\distB)$
\STATE $\xi\gets0$; $\hat\distA\gets0$; $\hat\distB\gets0$
\FOR{$j=j_{\min},\ldots,j_{\max}$}
 \STATE $C\gets \text{FFTconv}(\distA_j,\distB_j)+\text{FFTconv}(\distA_j,\hat\distB)+\text{FFTconv}(\hat\distA,\distB_j)$ \COMMENT{FFTs of length $4m$}
 \STATE $\xi\gets\xi+\proj(C)$
 \STATE $\hat\distA\gets\proj_{\spc{j+1}}(\hat\distA+\distA_j)$; $\hat\distB\gets\proj_{\spc{j+1}}(\hat\distB+\distB_j)$
\ENDFOR
\RETURN $\xi+\distA(0)\,\distB_++\distB(0)\,\distA_++\distA(0)\distB(0)\,\delta_0 +\bigl(1-(1-\distA(\infty))(1-\distB(\infty))\bigr)\delta_\infty$
\end{algorithmic}
\end{algorithm}

\begin{theorem}[Multiscale FFT merge]\label{thm:merge}
Let $\proj$ and $\proj_h$ for $h>0$ be projections onto $\grid$ and onto $h\mathbb Z_{\ge0}$ that satisfy properties \refTag{P1}--\refTag{P4}, let $\distA$ and $\distB$ be probability distributions on $\grid\cup\{\infty\}$ whose positive finite parts are supported on the bands $j_{\min},\ldots,j_{\max}$, and let $J\dfn j_{\max}-j_{\min}+1$. Then Algorithm~\ref{alg:ms-conv} returns $\proj(\distA*\distB)$, uses $O(Jm\log(m))$ arithmetic operations and $O(Jm)$ space, and the positive finite part of its output is supported on the bands $j_{\min},\ldots,j_{\max}+1$.
\end{theorem}

\begin{proof}[Proof outline]
The full proof appears in Appendix~\ref{apd:merge}. Every pair of positive finite values is counted in exactly one term of
\[
 \distA_+*\distB_+=\sum_{j=j_{\min}}^{j_{\max}}\bigl(\distA_j*\distB_j+\distA_j*\distB_{<j}+\distA_{<j}*\distB_j\bigr),
\]
namely, in the term of the larger of their bands. Since $\distA_j*\distB_{<j}$ is supported on $[2^j,\infty)$ and $\distA_j$ on $\spc j\mathbb Z_{\ge0}$, by the dyadic scaling,
\[
 \proj(\distA_j*\distB_{<j})\eqTag{P4}{=}\proj\bigl(\proj_{\spc j}(\distA_j*\distB_{<j})\bigr)\eqTag{P3}{=}\proj(\distA_j*\pref\distB j),
\]
and similarly for $\distA_{<j}*\distB_j$. By the linearity of the projection, this gives
\begin{equation}\label{eq:merge-identity}
 \begin{aligned}
 \proj(\distA*\distB)={}&\sum_{j=j_{\min}}^{j_{\max}}\proj\bigl(\distA_j*\distB_j+\distA_j*\pref\distB j+\pref\distA j*\distB_j\bigr)\\
 &+\distA(0)\,\distB_++\distB(0)\,\distA_++\distA(0)\distB(0)\,\delta_0+\bigl(1-(1-\distA(\infty))(1-\distB(\infty))\bigr)\delta_\infty.
 \end{aligned}
\end{equation}
Its second line collects the terms with the atoms at $0$ and at $\infty$.

By linearity and the projection hierarchy, $\pref\distA{j+1}=\proj_{\spc{j+1}}(\pref\distA j+\distA_j)$, which line~5 computes. Finally, since the projected prefixes are supported on $\{0,\spc j,\ldots,m\spc j\}$, in units of $\spc j$ the convolutions in line~3 are supported on the indices $m,\ldots,4m-2$, and thus FFTs of length $4m$ compute them exactly. Together with locality for the projections in lines~4 and~5, this gives the bounds on the running time, the space and the support of the output.
\end{proof}

\begin{remark}[Multiscale FFT merge with rounding]\label{rem:merge-rounding}
The projections that round every value up, or every value down, to a neighboring knot (Definition~\ref{def:projections}) also satisfy \refTag{P1}--\refTag{P4} (Lemma~\ref{lem:lattice}). Thus the multiscale FFT merge also computes the exact rounding of the sum, with the same bounds on its running time, its space and the support of its output (Theorem~\ref{thm:merge}). Rounding moves every value in one direction instead of preserving its mean, which preserves stochastic domination in that direction. For upper bounds, the sums are rounded up in the remove direction, whose loss is $\ln(\sumHat/t)$, and down in the add direction, whose loss is $-\ln(\sumHat/t)$, and the input is discretized in the matching directions. For lower bounds, which the CtD projection does not give, all these directions are reversed. Thus replacing the direct convolutions in the accountant of FS26 by multiscale FFT merges with the same rounding keeps the guarantees of FS26 (Remark~\ref{rem:rounding-guarantees}) and makes the running time nearly linear in $1/\alpha$. We use this accountant for comparisons in our experiments (Section~\ref{sec:experiments}).
\end{remark}

\section{Analysis}\label{sec:ctd}
\label{sec:guarantees}
We now state the guarantees of Algorithm~\ref{alg:ctd-alloc}.
\begin{theorem}[CtD accountant]\label{thm:ctd-discretized}
Let $\add L_t$ and $\rem L_t$ be the outputs of Algorithm~\ref{alg:ctd-alloc} on an input $L$, let $m$, $\indLo$ and $\indHi$ be the resolution and the lowest and the highest bands that it computes, let $J\dfn\indHi-\indLo+1$, and let $\gamma\dfn(\lceil\log_2(t)\rceil+1)\ln(1+1/m)$. Then $\add L_t$ and $\rem L_t$ are PLD realizations such that:
\begin{enumerate}
\item \textbf{Validity:} $\add L_t$ dominates $\psiAdd t(L)$, and $\rem L_t$ dominates $\psiRem t(\dual L)$.
\item \textbf{Tightness:} These dominations are $(\gamma,0)$-tight.
\item \textbf{$\chi^2$ error and bias:} If $m\ge\lceil\log_2(t)\rceil+1$, then in each direction in which the allocation PLD $\Psi$ that the output dominates satisfies $\chi^2(\Psi)<\infty$, the $\chi^2$ error of the output relative to $\Psi$ is between $0$ and $(\lceil\log_2(t)\rceil+1)(1+\chi^2(\Psi))/(4m^2)$. In each direction, the output and the allocation PLD that it dominates have the same mass at infinity, and the bias of the output relative to this PLD is between $0$ and $(\lceil\log_2(t)\rceil+1)\ln^2(1+1/m)/8$.
\item \textbf{Computational complexity:} The running time is $O((J+\log(t))\,m\log(m)\,\log(t))$, apart from lines~2 and~3, whose cost depends on the representation of the input, and the space is $O((J+\log(t))\,m)$.
\end{enumerate}
\end{theorem}
We outline the proofs of the first three parts below and give the full proof in Appendix~\ref{apd:ctd-thm}. The bounds on the running time and the space follow from Theorem~\ref{thm:merge}, since the positive finite part of every distribution in the computation is supported on at most $J+\lceil\log_2(t)\rceil+1$ bands (Claim~\ref{clm:bands}). The proofs do not use the choice of $m$ in line~1. Thus the theorem holds for every $m\ge2$.

\begin{remark}[Accountant with upper and lower rounding]\label{rem:rounding-guarantees}
The guarantees of Theorem~\ref{thm:ctd-discretized}, except for the bounds on the $\chi^2$ error and the bias, also hold for the accountant with rounding of Remark~\ref{rem:merge-rounding}, which computes each direction separately, as in FS26. Validity and tightness then hold with $\gamma$ replaced by $(\lceil\log_2(t-1)\rceil+2)\ln(1+1/m)$ in the remove direction. With the directions of the rounding reversed, the same holds for lower bounds, with each output dominated by the allocation PLD instead of dominating it, except that the outputs need not be PLD realizations. The outputs can still be used to provide valid lower bounds on the privacy profile as in FS26. Thus we also obtain a method to compute lower bounds in time that is nearly quadratically better than that of FS26.

The proofs are minor variants of the proof of Theorem~\ref{thm:ctd-discretized} and we omit them for brevity. Validity uses the monotonicity of rounding instead of the convex order, and tightness results from the grid resolution and the locality of the projection.
\end{remark}

\paragraph{The projection as a random map.} In the analysis we view the CtD projection $\proj$ onto $\grid$ as a random map that is applied independently of everything else and maps a value $x$ to a random knot $\tilde x$ with the distribution $\proj(\delta_x)$. The image $\widetilde Y$ of a random variable $Y$ then has the distribution $\proj(f_Y)$. The proofs of validity, tightness, and the bounds on the $\chi^2$ error and the bias use only the fact that the merges are exact (Theorem~\ref{thm:merge}), the grid resolution, and the locality \refTag{P1} and the mean preservation \refTag{P5} of the projection. For the random map they give $\tilde x\in\{u,v\}$ for every value $x$ between adjacent knots $u<v$, and $\expect{}{\tilde x}=x$ for every $x\ge0$.

\paragraph{Validity.}\label{sec:validity} Validity follows from the convex order, where $Y\cx Y'$ if $\expect{}{\varphi(Y)}\le\expect{}{\varphi(Y')}$ for every convex function $\varphi$ \citep{SS07}. By Jensen's inequality, $Y\cx Y'$ whenever $\expect{}{Y'\mid Y}=Y$, and two random variables that are ordered in the convex order have the same mean, since $x\mapsto x$ and $x\mapsto-x$ are convex.

\begin{lemma}[Validity]\label{lem:validity}
Let $L$ be a PLD realization for which $e^{-L}$ is bounded, let $\lrAvg t\dfn\frac1t\sum_{i=1}^te^{-L^{(i)}}$, and let $\lrHat$ be a bounded nonnegative random variable such that $\lrAvg t\cx\lrHat$. Then $-\ln(\lrHat)$ is a PLD realization that dominates $\psiAdd t(L)$, and its dual dominates $\psiRem t(\dual L)$.
\end{lemma}
\begin{proof}
Since $\lrAvg t\cx\lrHat$, they have the same mean $\expect{}{e^{-L}}\le1$, and since $\lrHat$ is bounded, $-\ln(\lrHat)$ is a PLD realization. By~\eqref{eq:profiles}, both privacy profiles are expectations of convex functions of the likelihood ratio, and thus the convex order between $\lrAvg t$ and $\lrHat$ gives domination in the add direction. Domination in the remove direction follows by~\eqref{eq:alloc-dual}, since the dual preserves domination (Claim~\ref{clm:dual-dom}).
\end{proof}

We apply Lemma~\ref{lem:validity} to the input $L$. We view the projection of the input in line~3 as projections at the leaves of the unfolded tree, and every projection as an independent random map (by Claim~\ref{clm:tree-dist}, this gives the distributions that repeated squaring computes). By the mean preservation of every projection, induction over the unfolded tree shows that the computed average $\lrHat\dfn\sumHat/t$ satisfies $\expect{}{\lrHat\mid\lrAvg t}=\lrAvg t$ for the exact average $\lrAvg t\dfn\frac1t\sum_{i=1}^te^{-L^{(i)}}$ (Lemma~\ref{lem:tree-coupling}). Thus $\lrAvg t\cx\lrHat$, and since $\add L_t=-\ln(\lrHat)$ and $\rem L_t=\dual{\add L_t}$, Lemma~\ref{lem:validity} gives the first part of Theorem~\ref{thm:ctd-discretized}.

This argument does not apply to the likelihood ratio under $P$. In the remove direction of FS26, the likelihood ratio $x$ under $P$ enters the privacy profile through the function $x\mapsto(1-e^\eps t/(x+s))_+$, where $s$ is the sum of the other factors. This function is strictly concave where it is positive, since its second derivative there is $-2e^\eps t/(x+s)^3$. Thus a mean-preserving projection of a value $x$ between two adjacent knots in this region strictly decreases the privacy profile.

\paragraph{Tightness.}\label{sec:tightness} Tightness bounds the privacy profiles of the output from above, and we obtain it from the same coupling. By locality of the projection and the grid resolution, every projection, including those at the leaves, changes a value by a factor of at most $1+1/m$, and thus $e^{-\gamma}\lrAvg t\le\lrHat\le e^{\gamma}\lrAvg t$, since the depth of the tree is $\lceil\log_2(t)\rceil$ (Claim~\ref{clm:depth} and Lemma~\ref{lem:tree-coupling}). In the add direction, the coupled losses $\psiAdd t(L)=-\ln(\lrAvg t)$ and $\add L_t=-\ln(\lrHat)$ are either both infinite or both finite, and in the latter case they differ by at most $\gamma$ (Lemma~\ref{lem:two-directions}). The tightness of the add direction does not necessarily imply the tightness of its dual. In the remove direction, we weight this coupling by $\lrHat$ and add the remaining mass $1-\expect{}{\lrAvg t}$ at $(+\infty,+\infty)$. Since $\expect{}{\lrHat\mid\lrAvg t}=\lrAvg t$, this gives a coupling of the remove losses, whose finite values again differ by at most $\gamma$ (Lemma~\ref{lem:two-directions}). These couplings give $\add L_t\domSt_{(\gamma,0)}\psiAdd t(L)$ and $\rem L_t\domSt_{(\gamma,0)}\psiRem t(\dual L)$. Together with validity, Claim~\ref{clm:stoch-hs} gives the second part of Theorem~\ref{thm:ctd-discretized}.

\paragraph{$\chi^2$ error and bias.}\label{sec:chi2} Rounding moves every value in the same direction, and thus increases the $\chi^2$ divergence (Claim~\ref{clm:chi2-stoch}) and the expected privacy loss by amounts of the first order in the spacing of the grid. In contrast, by locality of the projection and the grid resolution, a CtD projection changes a positive value by a factor of at most $1+1/m$, and together with mean preservation, for every nonnegative random variable $Y$ we get $\expect{}{\widetilde Y\mid Y}=Y$ and $(1+1/m)^{-1}Y\le\widetilde Y\le(1+1/m)Y$ (Claim~\ref{clm:random-map}). Thus the first-order effects of a projection cancel, and its $\chi^2$ error and bias are of the second order in the spacing, as are its effects on the mean and the variance of the privacy loss \citep[Section~6]{FS26clt}. In Appendices~\ref{apd:chi2-proof} and~\ref{apd:bias} we show that these errors add up over the at most $\lceil\log_2(t)\rceil+1$ projections on a path from a leaf to the root of the unfolded tree, which gives the bounds in the third part of Theorem~\ref{thm:ctd-discretized}.

\begin{remark}[Effect of the lower levels on $\eps$]\label{rem:lower-levels}
The loss parameter $\gamma$ of Theorem~\ref{thm:ctd-discretized} charges $\ln(1+1/m)$ to every level of projections. For all but the several highest levels this appears to be overly conservative. At height $j$, the CtD projections are applied to about $t/2^j$ independent sums of $2^j$ likelihood ratios, and these projected sums are then added up. If the higher levels added them exactly, the result would be random allocation over $t/2^j$ steps, with the distribution of the projected sums in place of that of the exact ones. The privacy losses of random allocation over many steps are approximately Gaussian, with variance equal to the $\chi^2$ divergence of the allocation PLD and mean equal to half of it \citep[Theorem~5.2]{FS26clt}. Thus the projections at such a level affect $\eps$ essentially only through the relative increase in this $\chi^2$ divergence, which is of the order of $\alpha^2/\chi^2(\Psi)$ for the allocation PLD $\Psi$. Only the several highest levels, at which few sums are added up, are not covered by this argument.
\end{remark}

\paragraph{Guarantees for a given accuracy.}\label{sec:end-to-end}
Substituting the resolution chosen in line~1 of Algorithm~\ref{alg:ctd-alloc} into Theorem~\ref{thm:ctd-discretized} gives the following guarantees in terms of $\alpha$ (Appendix~\ref{apd:input}).

\begin{corollary}[CtD accountant for a given accuracy]\label{cor:ctd-original}
Let $\alpha\in(0,1]$ and let $t\ge2$ be an integer. Let $\rem L$ and $\add L$ be PLD realizations with a total of $S$ atoms, and let $\wid$ be the width of the smallest interval that contains the finite values of $\rem L$ and of $-\add L$. There exists an algorithm that receives $\alpha$, $t$, $\rem L$ and $\add L$, and returns PLD realizations $\rem L_t$ and $\add L_t$ that $(\alpha,0)$-tightly dominate $\psiRem t(\rem L)$ and $\psiAdd t(\add L)$, in time
\[
 O\!\left(S+\frac{(\wid+\log(t))\log^2(t)}{\alpha}\log\!\left(\frac{\log(t)}{\alpha}\right)\right),
\]
and with space $O(S+(\wid+\log(t))\log(t)/\alpha)$. Moreover, in each direction the bias of the output relative to the allocation PLD $\Psi$ that it dominates is at most $\alpha^2/(8(\lceil\log_2(t)\rceil+1))$, and if $\chi^2(\Psi)<\infty$, then the $\chi^2$ error of the output relative to $\Psi$ is at most $\alpha^2(1+\chi^2(\Psi))/(4(\lceil\log_2(t)\rceil+1))$.
\end{corollary}

\begin{remark}[Truncation of unbounded inputs]\label{rem:continuous}
An input whose loss is unbounded, such as the PLD of the Gaussian mechanism, is first truncated, separately from the accountant. The truncation chooses a range of likelihood ratios that leaves out at most $\beta/(2t)$ of them under $Q$ and under $P$ on each side, projects the likelihood ratios below this range onto zero and its lower end by CtD, and moves those above it to its upper end, so that the truncated loss and its dual still dominate those of the input. Each factor of~\eqref{eq:psi} changes only on an event of probability at most $\beta/t$, and thus the truncation affects the final bounds only through the tail parameter $\beta$, which is negligible when $\beta$ is much smaller than the values of $\delta$ of interest. For the Gaussian mechanism with noise scale $\sigma$, the corresponding range of losses has width $1/\sigma^2+2\Phi^{-1}(1-\beta/(2t))/\sigma$, since $-L$ and $\dual L$ are distributed as $\mathcal N(\mp1/(2\sigma^2),1/\sigma^2)$, and the projection in line~3 needs only $O(Jm)$ evaluations of the standard normal CDF $\Phi$, since the probability and the first moment of the likelihood ratio on each cell are differences of two values of $\Phi$. Thus the accountant computes PLD realizations that $(\alpha,\beta)$-tightly dominate the allocation PLDs of the Gaussian mechanism using
\[
 O\!\left(\frac{\log^2(t)}{\alpha}\left(\log(t)+\frac1{\sigma^2}+\frac{\sqrt{\log(t/\beta)}}{\sigma}\right)\log\!\left(\frac{\log(t)}{\alpha}\right)\right)
\]
arithmetic operations and evaluations of $\Phi$.
\end{remark}

\section{Numerical Results}\label{sec:experiments}
We evaluate our accountant (Algorithm~\ref{alg:ctd-alloc}), referred to as {\em multiscale FFT}, using an implementation in the PLD accounting library \citep{PLD_acc_msfft}, which also contains the implementation of FS26. All the running times are of single-threaded runs on one core of a MacBook Pro M4 Max, and include the composition, when applicable.

\begin{figure}[t]
\centering
\includegraphics[width=\linewidth]{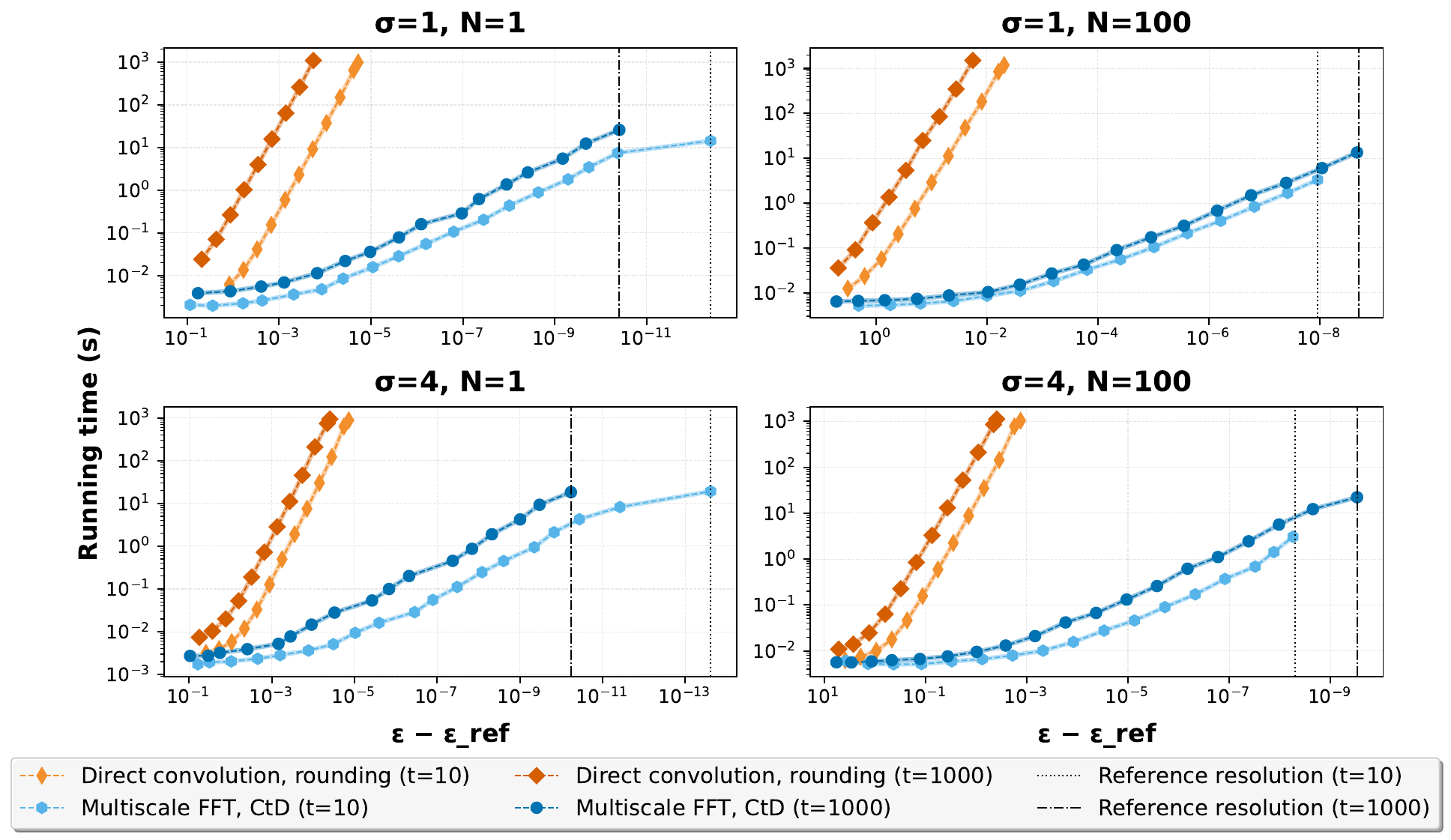}
\caption{Running time against the discrepancy $\eps-\eps_{\mathrm{ref}}$ of the computed upper bound $\eps$ at $\delta=10^{-6}$ from a reference $\eps_{\mathrm{ref}}$, for direct convolution with rounding and multiscale FFT with CtD projections, for $t\in\{10,1000\}$, a single allocation (left) and its $100$-fold composition (right), with $\sigma=1$ (top) and $\sigma=4$ (bottom). Here $\eps_{\mathrm{ref}}$ is the bound of multiscale FFT with at least twice the largest $m$ used in the same setting. The vertical lines mark the resolution of $\eps_{\mathrm{ref}}$ for each $t$, its difference from the bound with half its $m$. Each curve ends once its bound comes within twice this resolution of $\eps_{\mathrm{ref}}$. A bound within this resolution of $\eps_{\mathrm{ref}}$, or below it, is drawn on the line.}
\label{fig:runtime-accuracy}
\end{figure}

\paragraph{Running time and accuracy.}\label{sec:runtime-accuracy}
The two algorithms use their accuracy parameters differently, and CtD projections give only upper bounds. Therefore we compare the algorithms by the discrepancy of their upper bounds from a reference computed by multiscale FFT on a finer grid (Figure~\ref{fig:runtime-accuracy}). The regimes in the figure cover a wide range of resulting $\eps$ values at $\delta=10^{-6}$, from $\approx0.027$ (when $\sigma=4$ and $t=1000$, for a single allocation) to $\approx26$ (when $\sigma=1$ and $t=10$, for the $100$-fold composition).
At the finest accuracy of direct convolution, where a run takes about $1{,}000$ seconds, multiscale FFT reaches the same discrepancy in at most $0.03$ seconds, that is, it is faster by a factor of $10^4$ to $10^5$.
The running time grows roughly as the square of the inverse discrepancy for direct convolution and as its square root for multiscale FFT. Fitting $T=c+a\,x^k$, where $T$ is the running time and $x$ the inverse of the discrepancy, by least squares on $\ln(T)$ gives exponents $k$ between $1.96$ and $2.02$ for direct convolution and between $0.44$ and $0.55$ for multiscale FFT, over the points of the eight settings in Figure~\ref{fig:runtime-accuracy}.

\begin{figure}[t]
\centering
\includegraphics[width=\linewidth]{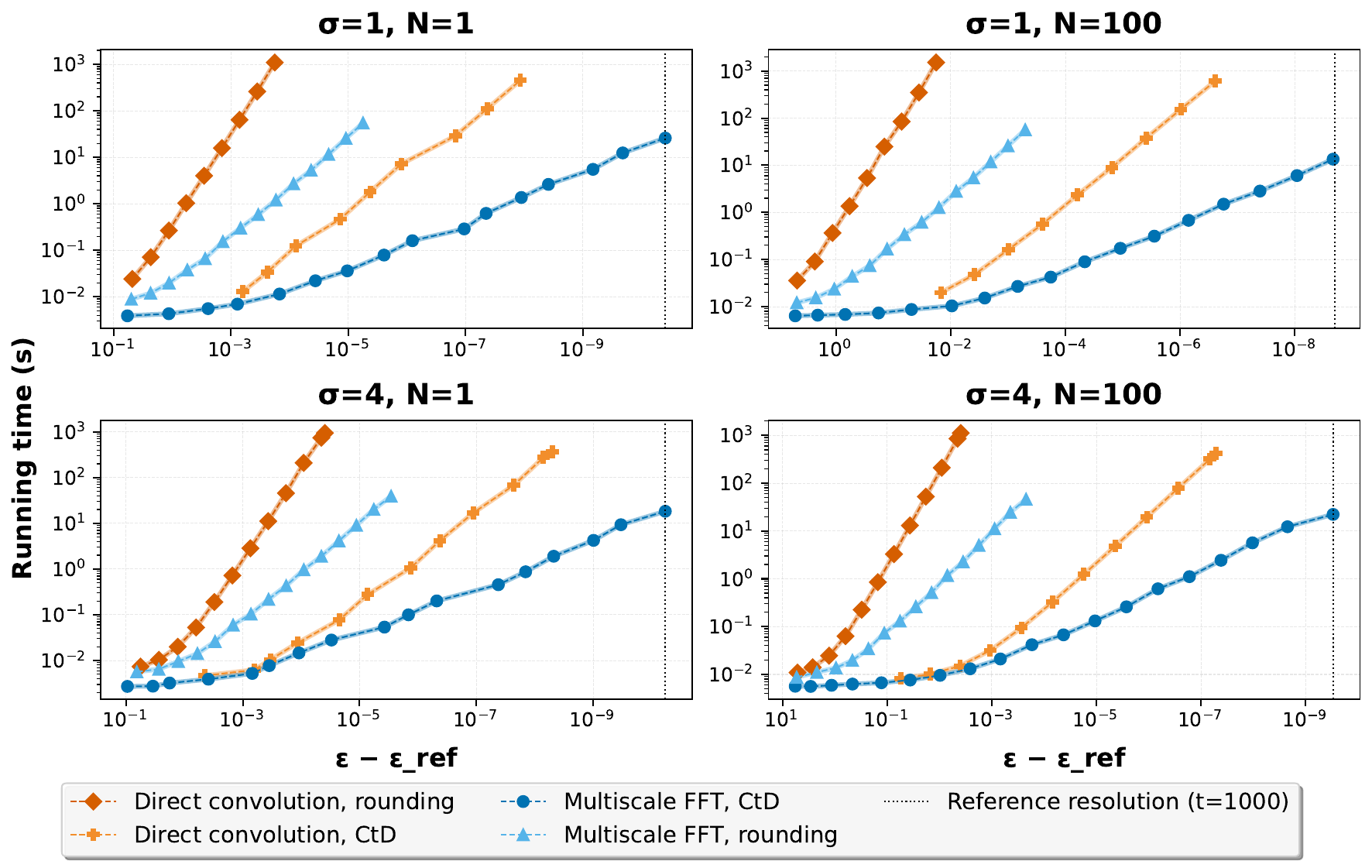}
\caption{The comparison of Figure~\ref{fig:runtime-accuracy} at $t=1000$ for the four combinations of the convolution method (direct convolution on the geometric grid or multiscale FFT on the dyadic grid) and the projection (rounding or CtD). The discrepancy $\eps-\eps_{\mathrm{ref}}$ is as in Figure~\ref{fig:runtime-accuracy}, and the dotted line marks the resolution of $\eps_{\mathrm{ref}}$, its difference from the bound with half its $m$.}
\label{fig:routes}
\end{figure}

\paragraph{Grid and projection.} These speedups correspond to the complete accountant, which changes both the grid and the projection. Figure~\ref{fig:routes} separates the two changes at $t=1000$ by combining each convolution method (direct on a geometric grid or multiscale-FFT on the dyadic) with each projection (stochastic domination rounding or hockey-stick domination via connect-the-dots). For each projection, the dyadic grid with multiscale FFT merges reaches the same or better accuracy faster than the geometric grid with direct convolution: at the finest accuracy of the geometric grid that the dyadic grid reaches, it is faster by two to three orders of magnitude.
For each grid, CtD projections reduce the discrepancy by orders of magnitude compared with rounding with the same accuracy parameters, and the difference grows with the number of compositions, which is consistent with the $\chi^2$ error of rounding accumulating over the compositions. Thus the grid and the merge determine the running time, and the projection determines the $\chi^2$ error. These routes also differ in other implementation choices, so this comparison indicates, but does not isolate, the contribution of each change.

\begin{figure}[t]
\centering
\includegraphics[width=0.8\linewidth]{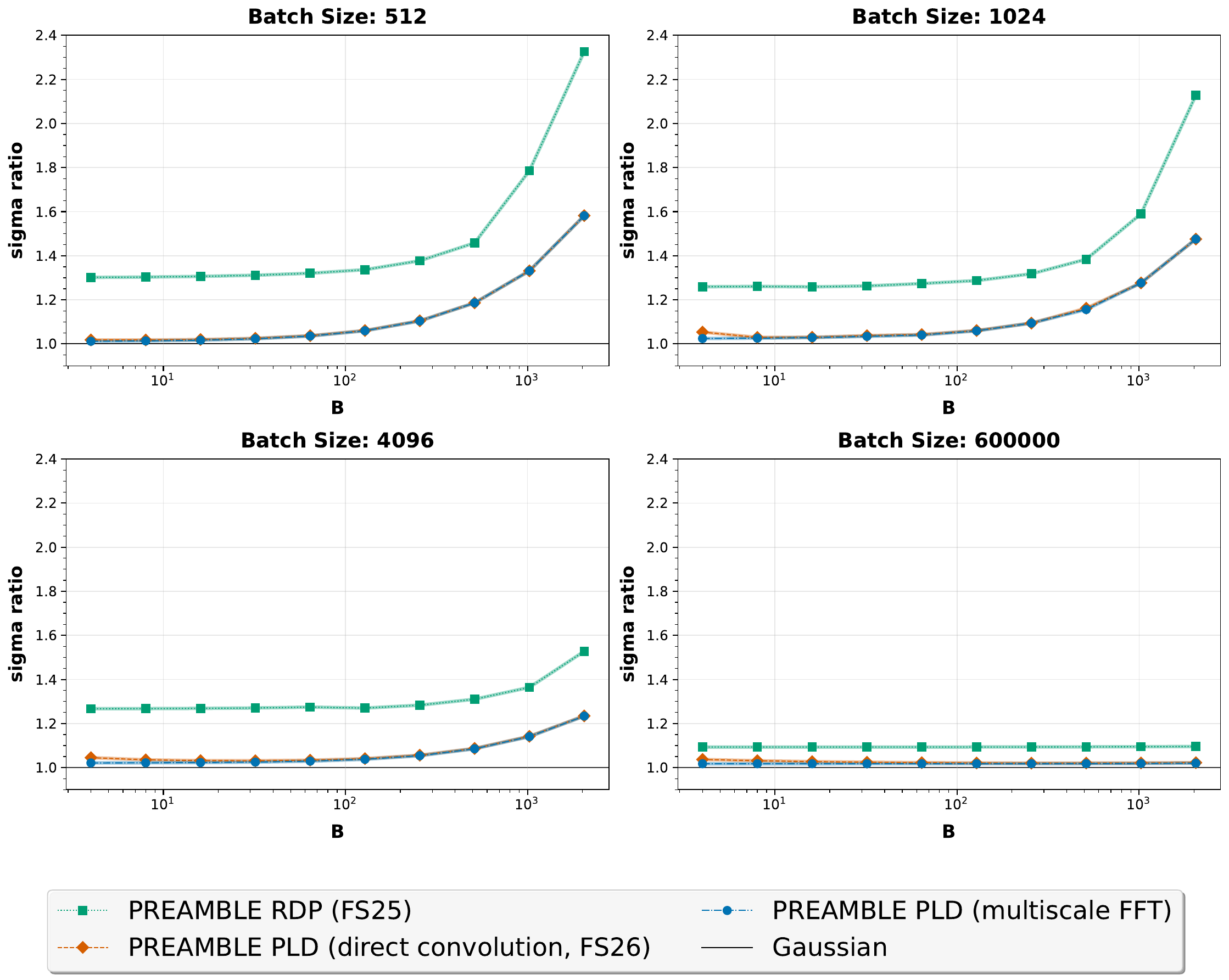}
\caption{The ratio between the noise scales required for $(1,10^{-6})$-DP by DP-SGD with PREAMBLE and by DP-SGD with Gaussian noise addition, as a function of the block size $B$, in the four settings of FS26. The results for RDP \citep{FS25ra} and for direct convolution are from FS26.}
\label{fig:preamble}
\end{figure}

\paragraph{DP-SGD with PREAMBLE.}\label{sec:preamble}
We follow the setting of FS26 for DP-SGD with Poisson subsampling of the users and PREAMBLE aggregation \citep{AFKRT26}, with $n=6\cdot10^5$ users, dimension $d=2^{20}$, communication budget $C=2^{15}$ and $10$ epochs.
Each selected user sends $k=C/B$ out of $t=d/B$ blocks of size $B$.
The analysis reduces to $1$-out-of-$32$ allocation composed $k$ times (up to $8192$ times for $B=4$), followed by Poisson subsampling and composition over the updates.
Since the allocation PLD is composed thousands of times, it has to be computed to a high accuracy.
For $B=4$ and expected batch size $1024$, the complete analysis at a fixed noise scale takes $0.06$ seconds with multiscale FFT and about an hour with direct convolution.
Accordingly, finding the noise scale that ensures $(1,10^{-6})$-DP takes at most a few minutes with multiscale FFT, instead of several hours with direct convolution.
Figure~\ref{fig:preamble} shows these noise scales, relative to that of DP-SGD in which each user sends the entire gradient, for the expected batch sizes $512$, $1024$, $4096$ and $600000$ and each $B$.
The resulting bounds are also slightly tighter: multiscale FFT never requires a larger noise scale than direct convolution, and requires a smaller one in $35$ of the $40$ settings, mostly for small $B$, where the number of inner compositions is large.
The example in the introduction uses $B=4$, expected batch size $1024$, the noise scale that direct convolution finds for $(1,10^{-6})$-DP and multiscale FFT with $m=1024$.
Multiscale FFT then computes the allocation PLD in $0.012$ seconds and gives the upper bound $0.92094$ on $\eps$ at $\delta=10^{-6}$, compared with $0.985$ for direct convolution.

The direct-convolution column is the run of FS26, with accuracy parameter $0.005$, tail parameter $10^{-8}$ and at most $10^6$ grid points.
FS26 used the expected batch size $1028$ instead of $1024$, so for batch size $1024$ we ran their implementation with the same configuration.
Multiscale FFT was run with accuracy parameter $0.01$, tail parameter $10^{-8}$ and at most $10^8$ points in the output grid.
After the division of the accuracy parameter among the compositions, the cap $m\le2^{18}$ binds for $B\le512$, that is, in $32$ of the $40$ settings.
In our runs, the allocation PLD is composed $k$ times by FFT, subsampled by the PLD subsampling of FS26, and composed over the updates by \texttt{dp\_accounting}.
The noise scale is found by a bisection that accepts any noise scale for which the bound on $\eps$ lies in $[0.98,1]$.
The running times at a fixed noise scale above were measured in the same way for both algorithms, whereas the search times are indicative only: the searches of multiscale FFT ran four at a time, and those of direct convolution are from its batch-$1024$ run.

\section{Discussion}\label{sec:discussion}
Multiscale FFT makes the running time of computing the allocation PLD nearly linear in the inverse accuracy, with the same $(\alpha,\beta)$ bound on the error of the privacy profiles as FS26. Empirically, the accuracy is even better: the error of the final bound on $\eps$ behaves essentially like the bias and the $\chi^2$ error, which are of the second order in the spacing of the grid (Section~\ref{sec:runtime-accuracy}). As we argue in Remark~\ref{rem:lower-levels}, the projections at all but the several highest levels of the unfolded tree affect $\eps$ essentially only through the relative increase in the $\chi^2$ divergence. A natural question is whether this also holds for the highest levels, at which few independent sums are projected.

\subsection*{AI Usage}
The authors formulated the problem and proposed the two key elements of the algorithm. The development of these ideas and writing were performed through an iterative process in which AI (Claude Opus 5.5 and ChatGPT Astra 6) generated the implementation, details of proofs and wrote parts of the prose based on author-provided guidance and structure. The resulting text was also extensively edited and proofread by the authors. The implementation of the algorithm and evaluation code was written by AI under close supervision and auditing by the authors. The authors take full responsibility for the content of this paper.

\subsection*{Acknowledgments}
Shenfeld's work was supported in part by ERC grant 101125913, Simons Foundation Collaboration 733792, Israel Science Foundation (ISF) grant 2861/20, and Apple. Views and opinions expressed are however those of the authors only and do not necessarily reflect those of the European Union or the European Research Council Executive Agency. Neither the European Union nor the granting authority can be held responsible for them.

\clearpage
\appendix

\section{Preliminary Statements}\label{apd:prelims}
In this section we give the statements from FS26 and ZDW22 that we use, following their notation, together with the proof of~\eqref{eq:alloc-dual}.

\subsection{Domination}
The pair $(\mathcal N(1,\sigma^2),\mathcal N(0,\sigma^2))$ dominates the Gaussian mechanism with sensitivity one and noise scale $\sigma$ in both directions, and its PLD is $\mathcal N(1/(2\sigma^2),1/\sigma^2)$ \citep{ZDW22}. We say that a pair $(P,Q)$ {\em dominates} a $t$-step algorithm $\alg$ if it is a dominating pair in the remove direction for every step of $\alg$ and every view (in the terms of \citep[Definition~2.6]{FS26pld}, $\alg$ is dominated by a randomizer with output distributions $P$ and $Q$).

\begin{lemma}[{\cite[Lemmas~2.7 and~2.8 and Claim~2.9]{FS26pld}}]\label{lem:reduction}
Let $(P,Q)$ dominate a $t$-step algorithm $\alg$. Then $\alloc t\alg$ is dominated by $(\bar P_t,Q^t)$ in the remove direction and by $(Q^t,\bar P_t)$ in the add direction. Moreover, $\alloc{t,k}{\alg}$ is dominated by $(\bar P_{\lfloor t/k\rfloor}^{\,m_f}\times\bar P_{\lceil t/k\rceil}^{\,m_c},Q^t)$ in the remove direction and by the reversed pair in the add direction, where $m_f+m_c=k$, $m_f\lfloor t/k\rfloor+m_c\lceil t/k\rceil=t$, and powers denote product distributions.
\end{lemma}

\begin{claim}[{\cite[Section~3 and Claims~A.1 and~A.2]{FS26pld}}]\label{clm:pld-real}
Every PLD $\lossRVfunc PQ$ is a PLD realization, and its dual is $\lossRVfunc QP$. Conversely, a PLD realization $L$ is the PLD of the pair consisting of the distribution of $L$ and the distribution of $-\dual L$, and $\dual L$ is a PLD realization and the PLD of the reversed pair.
\end{claim}

\begin{claim}[{\cite[Lemma 28]{ZDW22}}]\label{clm:dual-dom}
If a PLD realization $\upBnd$ dominates a PLD realization $L$, then $\dual\upBnd$ dominates $\dual L$.
\end{claim}

\begin{claim}[{\cite[Claim~4.2]{FS26pld}}]\label{clm:stoch-hs}
If $\lowBnd\domSt_{(\alpha,\beta)}\upBnd$ then $\HockeyStickFunc{e^\eps}{\lowBnd}\le\HockeyStickFunc{e^{\eps-\alpha}}{\upBnd}+\beta$ for every $\eps$. In particular, stochastic domination implies domination.
\end{claim}

\subsection{Random Allocation and Duality}\label{apd:represent}
\begin{theorem}[{\cite[Theorem~4.4]{FS26pld}}]\label{thm:psi}
For a PLD realization $L$, let $\psiRem t(L)$ and $\psiAdd t(L)$ be as in~\eqref{eq:psi}. Then for any two distributions $P$ and $Q$, $\lossRVfunc{\bar P_t}{Q^t}=\psiRem t(\lossRVfunc PQ)$ and $\lossRVfunc{Q^t}{\bar P_t}=\psiAdd t(\lossRVfunc QP)$.
\end{theorem}

\begin{lemma}[{\cite[Lemma~4.5]{FS26pld}}]\label{lem:mult-tight}
Let $V_1,\ldots,V_n$ be independent, let $U_1,\ldots,U_n$ be independent, and let $V_i\domSt_{(\alpha,\beta_i)}U_i$ for every $i$. Then $\ln(\sum_ie^{V_i})\domSt_{(\alpha,\sum_i\beta_i)}\ln(\sum_ie^{U_i})$.
\end{lemma}
FS26 state this lemma for two summands, and the general case follows by induction. The same proof applies when $\alpha=0$ or $\beta_i=0$. For $\alpha=\beta_i=0$ it states that stochastic domination is preserved by $\ln(\sum_ie^{V_i})$. The normalization by $t$ in~\eqref{eq:psi} is a common shift.

\begin{claim}\label{clm:alloc-dual}
For every PLD realization $L$ and integer $t\ge1$, $\psiRem t(\dual L)=\dual{\psiAdd t(L)}$.
\end{claim}
\begin{proof}
By Claim~\ref{clm:pld-real}, $L$ is the PLD of a pair of distributions, that is, $L=\lossRVfunc QP$ for some $P$ and $Q$, and then $\dual L=\lossRVfunc PQ$. By Theorem~\ref{thm:psi}, $\psiAdd t(L)=\lossRVfunc{Q^t}{\bar P_t}$ and $\psiRem t(\dual L)=\lossRVfunc{\bar P_t}{Q^t}$, which is the dual of $\lossRVfunc{Q^t}{\bar P_t}$.
\end{proof}

\section{\texorpdfstring{Missing Details from Section~\ref{sec:multiscale}}{Missing Details from Section 3}}\label{apd:ms}

\subsection{Projections and Their Lattice Identities}\label{apd:lattice}
\begin{definition}[Projections]\label{def:projections}
Let $S\subseteq[0,\infty)$ be a set of knots that contains $0$ and has no accumulation point in $(0,\infty)$, such as $\grid$ or the lattice $h\mathbb Z_{\ge0}$ for $h>0$. For $x\in[0,\sup S]$, let $u\le x\le v$ be the largest knot that is at most $x$ and the smallest knot that is at least $x$ (so that $u=v=x$ if $x\in S$). For $u<v$ we define
\[
 \proj^\uparrow_S(\delta_x)\dfn\delta_v,
 \qquad
 \proj^\downarrow_S(\delta_x)\dfn\delta_u,
 \qquad
 \proj^\ctd_S(\delta_x)\dfn\frac{v-x}{v-u}\,\delta_u+\frac{x-u}{v-u}\,\delta_v,
\]
and for $x\in S$ all three are equal to $\delta_x$. We extend each projection linearly to finite measures on $[0,\sup S]$, and to measures with an atom at $+\infty$ by leaving that atom unchanged.
\end{definition}

\begin{claim}\label{clm:grid}
For every integer $m\ge2$, the grid $\grid$ satisfies \refTag{G1} and \refTag{G2}.
\end{claim}
\begin{proof}
Let $x>0$ and let $j\in\mathbb Z$ be such that $x\in\band j$. Then $u\le x\le v$ for two consecutive knots $u<v$ with $u\in\band j$ and $v=u+\spc j$, also when $v=2^{j+1}$. Since $u\ge2^j=m\spc j$, we have $v/u=1+\spc j/u\le1+1/m$, which proves \refTag{G1}. For \refTag{G2}, every knot that is at least $2^j$ is of the form $2^{j'}(1+i/m)$ with $j'\ge j$ and $0\le i\le m-1$, and it is equal to $2^{j'-j}(m+i)\spc j$.
\end{proof}

\begin{lemma}\label{lem:lattice}
Given $\rho\in\{\uparrow,\downarrow,\ctd\}$, the maps $\proj\dfn\proj^\rho_{\grid}$ and $\proj_h\dfn\proj^\rho_{h\mathbb Z_{\ge0}}$ for $h>0$ are projections in the sense of Section~\ref{sec:ctd-projection}, and for every $h>0$, $j\in\mathbb Z$ and all finite measures $\xi$ and $\zeta$, each of these projections $\proj_S$ satisfies: $\proj_S(\delta_x)$ is supported on $\{u,v\}$ for every $x$ between adjacent knots $u<v$ of $S$ \refTag{P1}, $\proj_S(\xi+\zeta)=\proj_S(\xi)+\proj_S(\zeta)$ \refTag{P2}, $\proj_h(\xi*\zeta)=\proj_h(\xi)*\zeta$ if $\zeta$ is supported on $h\mathbb Z_{\ge0}$ \refTag{P3}, and $\proj_{2h}(\proj_h(\xi))=\proj_{2h}(\xi)$ and $\proj(\proj_{\spc j}(\xi))=\proj(\xi)$ if $\xi$ is supported on $[2^j,\infty)$ \refTag{P4}. Moreover, for $\rho=\ctd$, $\proj_S(\delta_x)$ has mean $x$ for every $x\ge0$ \refTag{P5}.
\end{lemma}
The second part of \refTag{P4} does not hold for measures that are not supported on $[2^j,\infty)$, since below $2^j$ the knots of $\grid$ are finer than the lattice $\spc j\mathbb Z$.

\begin{proof}
Each map $\proj^\rho_S$ sends a finite measure to a measure on $S\cup\{\infty\}$ and fixes the knots and $+\infty$. Since it is extended linearly, it satisfies \refTag{P2} and fixes every measure supported on $S\cup\{\infty\}$, that is, it is a projection. By definition, it satisfies \refTag{P1}. For $\rho=\ctd$ and $x$ between adjacent knots $u<v$ of $S$, the mean of $\proj_S(\delta_x)$ is $u(v-x)/(v-u)+v(x-u)/(v-u)=x$, which is \refTag{P5}. For a set of knots $S$, a knot $r\in S$ and a projection $\rho$, let $k^\rho_S(x,r)\dfn\proj^\rho_S(\delta_x)(\{r\})$ denote the mass that the projection of $x$ assigns to $r$. As a function of $x$, $k^\ctd_S(\cdot,r)$ is affine between every two adjacent knots of $S$, $k^\uparrow_S(\cdot,r)$ is constant on every interval $(s,s']$ between adjacent knots, and $k^\downarrow_S(\cdot,r)$ is constant on every interval $[s,s')$ between adjacent knots.

We first show that if $T\subseteq S$ are two sets of knots then $\proj_T(\proj_S(\xi))=\proj_T(\xi)$ for every finite measure $\xi$ on which both projections are defined. By \refTag{P2} it suffices to consider $\xi=\delta_x$. If $x\in S$ the claim is trivial. Otherwise, let $x$ lie between adjacent knots $u<v$ of $S$. Since $T\subseteq S$, the interval $[u,v]$ is contained in an interval between adjacent knots of $T$. For $\rho=\ctd$ and $r\in T$,
\[
 \proj_T(\proj_S(\delta_x))(\{r\})=\frac{v-x}{v-u}k_T(u,r)+\frac{x-u}{v-u}k_T(v,r)=k_T(x,r),
\]
since $k_T(\cdot,r)$ is affine on $[u,v]$. For $\rho={\uparrow}$ we have $\proj_S(\delta_x)=\delta_v$ with $x\in(u,v]$, and $k_T(v,r)=k_T(x,r)$ since $k_T(\cdot,r)$ is constant on an interval that contains $(u,v]$. The case $\rho={\downarrow}$ is symmetric.

The first part of \refTag{P4} follows by taking $T=2h\mathbb Z_{\ge0}\subseteq S=h\mathbb Z_{\ge0}$. For its second part, let $T\dfn\grid\cap[2^j,\infty)$ and $S\dfn \spc j\mathbb Z\cap[2^j,\infty)$, so that $T\subseteq S$ by \refTag{G2}. Since $2^j\in \spc j\mathbb Z$, the projection onto $\spc j\mathbb Z_{\ge0}$ of a value that is at least $2^j$ is supported on $[2^j,\infty)$ and is equal to its projection onto $S$. Similarly, since $2^j\in\grid$, the projection onto $\grid$ of a measure supported on $[2^j,\infty)$ is equal to its projection onto $T$. Thus $\proj(\proj_{\spc j}(\xi))=\proj_T(\proj_S(\xi))=\proj_T(\xi)=\proj(\xi)$. For \refTag{P3}, for $y\in h\mathbb Z_{\ge0}$ the adjacent knots of $h\mathbb Z_{\ge0}$ around $x+y$ are the adjacent knots around $x$ translated by $y$, and the weights (for CtD) or the choice of the knot (for rounding) are the same. Thus $\proj_h(\delta_{x+y})=\proj_h(\delta_x)*\delta_y$, and integrating over $\xi(dx)\,\zeta(dy)$ gives the claim. Atoms at $+\infty$ are not affected by any of the operations.
\end{proof}

\subsection{The Multiscale FFT Merge}\label{apd:merge}
\begin{proof}[Proof of Theorem~\ref{thm:merge}]
\emph{The identity.} Writing $\distA=\distA(0)\,\delta_0+\distA_++\distA(\infty)\,\delta_\infty$ and similarly for $\distB$, we get $\distA*\distB=\distA_+*\distB_++\rho$, where $\rho$ denotes the second line of~\eqref{eq:merge-identity}. Since $\distA_+$ and $\distB_+$ are supported on $\grid$, the measure $\rho$ is supported on $\grid\cup\{\infty\}$, and thus $\proj(\rho)=\rho$. Every pair of values from the positive bands $i$ and $k$ is counted in exactly one term of
\begin{equation}\label{eq:scale-decomposition}
 \distA_+*\distB_+=\sum_{j=j_{\min}}^{j_{\max}}\bigl(\distA_j*\distB_j+\distA_j*\distB_{<j}+\distA_{<j}*\distB_j\bigr),
\end{equation}
namely, in the term with $j=\max\{i,k\}$. For every $j$, the measure $\distA_j*\distB_{<j}$ is supported on $[2^j,\infty)$, and $\distA_j$ is supported on $\spc j\mathbb Z_{\ge0}$ by \refTag{G2}. Thus \refTag{P4} and \refTag{P3} with $\zeta=\distA_j$ give
\[
 \proj(\distA_j*\distB_{<j})
 \eqTag{P4}{=}\proj\bigl(\proj_{\spc j}(\distA_j*\distB_{<j})\bigr)
 \eqTag{P3}{=}\proj\bigl(\distA_j*\proj_{\spc j}(\distB_{<j})\bigr)
 =\proj(\distA_j*\pref\distB j),
\]
and similarly $\proj(\distA_{<j}*\distB_j)=\proj(\pref\distA j*\distB_j)$. Therefore
\begin{align*}
 \proj(\distA*\distB)
 &\eqTag{P2}{=}\proj(\distA_+*\distB_+)+\proj(\rho)\\
 &\eqExp{a}{=}\sum_{j=j_{\min}}^{j_{\max}}\bigl(\proj(\distA_j*\distB_j)+\proj(\distA_j*\distB_{<j})+\proj(\distA_{<j}*\distB_j)\bigr)+\rho\\
 &\eqExp{b}{=}\sum_{j=j_{\min}}^{j_{\max}}\bigl(\proj(\distA_j*\distB_j)+\proj(\distA_j*\pref\distB j)+\proj(\pref\distA j*\distB_j)\bigr)+\rho\\
 &\eqTag{P2}{=}\sum_{j=j_{\min}}^{j_{\max}}\proj\bigl(\distA_j*\distB_j+\distA_j*\pref\distB j+\pref\distA j*\distB_j\bigr)+\rho,
\end{align*}
where (a) follows from~\eqref{eq:scale-decomposition}, \refTag{P2} and $\proj(\rho)=\rho$, and (b) from the two identities above. This proves~\eqref{eq:merge-identity}.

\emph{The algorithm.} We show by induction on $j$ that $\hat\distA=\pref\distA j$ and $\hat\distB=\pref\distB j$ at the beginning of iteration $j$ of Algorithm~\ref{alg:ms-conv}. For $j=j_{\min}$ we have $\distA_{<j}=0$, and thus $\pref\distA j=\proj_{\spc j}(0)=0=\hat\distA$, since a projection fixes the zero measure. If $\hat\distA=\pref\distA j$ at the beginning of iteration $j$, then line~5 sets $\hat\distA$ to
\begin{align*}
 \proj_{\spc{j+1}}(\pref\distA j+\distA_j)
 &\eqTag{P2}{=}\proj_{\spc{j+1}}\bigl(\proj_{\spc j}(\distA_{<j})\bigr)+\proj_{\spc{j+1}}(\distA_j)\\
 &\eqTag{P4}{=}\proj_{\spc{j+1}}(\distA_{<j})+\proj_{\spc{j+1}}(\distA_j)
 \eqTag{P2}{=}\proj_{\spc{j+1}}(\distA_{<j+1})
 =\pref\distA{j+1},
\end{align*}
where \refTag{P4} applies since $\spc{j+1}=2\spc j$, and the last two steps use $\distA_{<j+1}=\distA_{<j}+\distA_j$. The same holds for $\hat\distB$. Thus in iteration $j$, line~3 computes the argument of the $j$th projection in~\eqref{eq:merge-identity}, and line~4 adds this projection to $\xi$. Therefore Algorithm~\ref{alg:ms-conv} returns the right-hand side of~\eqref{eq:merge-identity}, which is equal to $\proj(\distA*\distB)$ by the identity.

\emph{Running time, space and support.} By \refTag{P1}, each projection moves a value between adjacent knots to one or both of these knots. In iteration $j$ we represent measures supported on $\spc j\mathbb Z\cap[0,4m\spc j)$ by the arrays of their masses at the indices $0,\ldots,4m-1$ in units of $\spc j$. By the definition of $\grid$, the knots in band $j$ are the indices $m,\ldots,2m-1$, and those in band $j+1$ are the even indices $2m,\ldots,4m-2$, since $\spc{j+1}=2\spc j$. Since $m\spc j=2^j$ is a knot of $\spc j\mathbb Z_{\ge0}$, the projection $\pref\distA j$ of the values of $\distA$ in $(0,2^j)$ is supported on the indices $0,\ldots,m$, and so is $\pref\distB j$. By the previous part, these are the values of $\hat\distA$ and $\hat\distB$ in iteration $j$. Thus the measure $C$ in line~3 is supported on the indices $m,\ldots,4m-2$: the term $\distA_j*\distB_j$ on the indices $2m,\ldots,4m-2$, and the other two terms on the indices $m,\ldots,3m-1$. Since $4m-2<4m$, the cyclic convolutions of length $4m$ of the zero-padded arrays are equal to the linear ones, and thus $C$ is computed by a constant number of FFTs of length $4m$, that is, in $O(m\log(m))$ operations. The masses of $C$ at the indices $m,\ldots,2m-1$ and at the even indices $2m,\ldots,4m-2$ lie on knots of $\grid$ and are fixed by $\proj$, and every odd index $i$ with $2m<i<4m-2$ lies between the adjacent knots $i-1$ and $i+1$. Thus line~4 obtains $\proj(C)$ by moving (for rounding) or splitting (for CtD) the mass at each such index to these two indices, in $O(m)$ operations, and $\proj(C)$ is supported on the bands $j$ and $j+1$. Similarly, $\hat\distA+\distA_j$ is supported on the indices $0,\ldots,2m-1$, and line~5 obtains its projection onto $\spc{j+1}\mathbb Z_{\ge0}$, that is, onto the even indices, by moving or splitting the mass at each odd index to its two neighbors, in $O(m)$ operations. The result is supported on the indices $0,\ldots,m$ in units of $\spc{j+1}$, as needed for the next iteration. Line~7 takes $O(Jm)$ operations, since $\distA_+$ and $\distB_+$ are supported on the $Jm$ knots of $J$ bands. Summing over the $J$ iterations, Algorithm~\ref{alg:ms-conv} uses $O(Jm\log(m))$ operations. The positive finite part of its output is the sum of the measures $\proj(C)$ over the $J$ iterations and of $\distA(0)\,\distB_++\distB(0)\,\distA_+$, and thus it is supported on the bands $j_{\min},\ldots,j_{\max}+1$. Finally, the inputs, $\xi$ and the output are supported on at most $(J+1)m+2$ points, and the prefixes and the buffers of the FFT take $O(m)$ space, which gives the space bound $O(Jm)$.
\end{proof}

\subsection{The Unfolded Tree of Repeated Squaring}\label{apd:depth}
By Lemma~\ref{lem:mult-tight}, the loss parameter of a sum is the larger of the loss parameters of its summands. Thus the errors of the merges accumulate along the {\em depth} of the unfolded tree (Section~\ref{sec:ctd-alloc}) rather than over all of its merges. The same observation applies to the accountant of FS26, whose analysis charges its spacing $\lossPar$ for the discretization of the input and for every merge. Its remove direction merges the sum of $t-1$ copies with one more factor, so its unfolded tree has depth $\lceil\log_2(t-1)\rceil+1\le\lceil\log_2(t)\rceil+1$, and its guarantees also hold with $\lossPar=\alpha/(\lceil\log_2(t)\rceil+2)$ in place of $\alpha/(2\lceil\log_2(t)\rceil+1)$.

\begin{claim}\label{clm:depth}
For $n\ge1$, repeated squaring for the sum of $n$ copies uses at most $2\lceil\log_2(n)\rceil$ merges, and its unfolded tree has depth $\lceil\log_2(n)\rceil$.
\end{claim}
\begin{proof}
The case $n=1$ is trivial, so let $n\ge2$ and let $n=\sum_{k=1}^s2^{b_k}$ with $b_1<\cdots<b_s$, so that $b_s=\lfloor\log_2(n)\rfloor$. Repeated squaring performs $b_s$ squarings and $s-1$ further merges. Since $b_1,\ldots,b_s$ are distinct integers in $\{0,\ldots,b_s\}$, we have $s\le b_s+1$, and thus $b_s+s-1\le2\lfloor\log_2(n)\rfloor\le2\lceil\log_2(n)\rceil$. Repeated squaring computes the sum of $2^b$ copies for $b\le b_s$ by successive merges of a sum with itself. Thus the unfolded tree of the sum of $2^b$ copies is a complete binary tree of depth $b$. The accumulated sum is initialized to the sum of $2^{b_1}$ copies (of depth $b_1$) and, for $k=2,\ldots,s$, it is merged with the sum of $2^{b_k}$ copies. After this merge its depth is $\max\{D_{k-1},b_k\}+1=b_k+1$, where $D_{k-1}\le b_{k-1}+1\le b_k$ is its depth before the merge. Thus the depth of the result is $b_s=\log_2(n)$ if $s=1$ and $b_s+1=\lfloor\log_2(n)\rfloor+1$ otherwise, which is $\lceil\log_2(n)\rceil$ in both cases.
\end{proof}

The next claim justifies viewing every projection in the unfolded tree as an independent random map, as the analysis in Section~\ref{sec:ctd} does.

\begin{claim}\label{clm:tree-dist}
Let $\distA$ be a probability distribution on $\grid$ whose positive values lie in finitely many bands, and let $n\ge1$. Consider the unfolded tree of repeated squaring for the sum of $n$ copies, whose leaves are independent random variables with distribution $\distA$, and in which every internal node applies the CtD projection onto $\grid$, as an independent random map, to the sum of the values of its children. Then the value at every node has the distribution that repeated squaring computes at the corresponding merge, and in particular the value at the root has the distribution $\selfConv(\distA,n)$.
\end{claim}
\begin{proof}
We use induction on the height of the node. A leaf has the distribution $\distA=\distA^{(0)}$. Consider an internal node whose children have, by the induction hypothesis, the distributions $\xi_1$ and $\xi_2$ computed by repeated squaring. The subtrees of the two children have disjoint sets of leaves and of projections. Thus the values at the children are independent, and their sum has the distribution $\xi_1*\xi_2$. The image of an independent random variable with distribution $f$ under the random map has the distribution $\proj(f)$ (Claim~\ref{clm:random-map}). Therefore the value at the node has the distribution $\proj(\xi_1*\xi_2)$, which is equal to $\msConv(\xi_1,\xi_2)$ by Theorem~\ref{thm:merge} and Lemma~\ref{lem:lattice}, that is, to the distribution computed at the corresponding merge.
\end{proof}

\section{\texorpdfstring{Missing Details from Section~\ref{sec:ctd}}{Missing Details from Section 4}}\label{apd:ctd-proofs}

\subsection{The CtD Projection and the Pessimistic CtD Approximation}\label{apd:ctd-pessimistic}
We first prove the properties of the CtD projection as a random map that Section~\ref{sec:guarantees} states.

\begin{claim}\label{clm:random-map}
Let $\tilde x$ denote the image of a value $x\ge0$ under the CtD projection onto $\grid$, viewed as a random map. Then $\tilde x$ has the distribution $\proj(\delta_x)$, and the image of a random variable with distribution $f$ under an independent random map has the distribution $\proj(f)$. Moreover, for every nonnegative random variable $Y$ and its image $\widetilde Y$ under an independent random map,
\begin{equation}\label{eq:projection-coupling}
 \expect{}{\widetilde Y\mid Y}=Y,
 \qquad
 \Bigl(1+\frac1m\Bigr)^{-1}Y\le\widetilde Y\le\Bigl(1+\frac1m\Bigr)Y,
\end{equation}
and in particular $\prob{}{\widetilde Y=0}=\prob{}{Y=0}$ and $\expect{}{\widetilde Y}=\expect{}{Y}$.
\end{claim}
\begin{proof}
If $x$ is a knot then $\tilde x=x$. Otherwise, let $u<x<v$ be the adjacent knots around $x$. The map moves $x$ to $v$ with probability $(x-u)/(v-u)$ and to $u$ otherwise, which gives the distribution $\proj(\delta_x)$. The distribution of the image of a random variable with distribution $f$ is $\int\proj(\delta_x)\,f(dx)$, which is $\proj(f)$ by the linear extension of $\proj$. Since $\tilde x\in\{u,v\}$ by \refTag{P1} and $v\le(1+1/m)u$ by \refTag{G1} (Claim~\ref{clm:grid}), we have $x/(1+1/m)\le\tilde x\le(1+1/m)x$ for every $x\ge0$. Conditioning on $Y$ and applying this bound and \refTag{P5} with $x=Y$ gives~\eqref{eq:projection-coupling}. In particular, $\widetilde Y=0$ if and only if $Y=0$, and taking expectations in the first identity gives $\expect{}{\widetilde Y}=\expect{}{Y}$.
\end{proof}

Next we show that the CtD projection of likelihood ratios gives the pessimistic CtD approximation of \citet{DGKKM22}. Given a PLD and a grid of losses, this approximation is the PLD supported on the grid and on $+\infty$ whose privacy profile, as a function of $e^\eps$, is equal to that of the given PLD at the points of the grid and is affine between them.

\begin{claim}\label{clm:ctd-pessimistic}
Let $S\subseteq[0,\infty)$ be a set of knots that contains $0$, let $L$ be a PLD realization such that $e^{-L}\le s_{\max}$ for a knot $s_{\max}$ of $S$, and let $\widetilde L$ be the PLD realization for which $e^{-\widetilde L}$ is the CtD projection of $e^{-L}$ onto $S$. Then, as functions of $e^\eps$:
\begin{enumerate}
\item The privacy profile of $\dual{\widetilde L}$ is equal to that of $\dual L$ at every point $e^\eps\in S$, is affine between every two consecutive points of $S$ (including $0$, where both profiles tend to $1$), and is constant for $e^\eps\ge s_{\max}$.
\item The privacy profile of $\widetilde L$ is equal to that of $L$ at every point $e^\eps=1/s$ with $s\in S\setminus\{0\}$, is affine between every two such consecutive points and for $e^\eps\le1/s_{\max}$, and is constant for $e^\eps\ge1/s_{\min}$ if $S$ has a smallest positive knot $s_{\min}$.
\end{enumerate}
Thus $\dual{\widetilde L}$ and $\widetilde L$ are the pessimistic CtD approximations of $\dual L$ and of $L$ on the grids of losses $\ln(S\setminus\{0\})$ and $-\ln(S\setminus\{0\})$, respectively.
\end{claim}
\begin{proof}
Let $\lr\dfn e^{-L}$ and $\widetilde\lr\dfn e^{-\widetilde L}$, so that $\expect{}{\widetilde\lr}=\expect{}{\lr}$. Let $C(s)\dfn1-\expect{}{\lr}+\expect{}{(\lr-s)_+}$, and let $\widetilde C$ be defined in the same way for $\widetilde\lr$. By~\eqref{eq:profiles}, the privacy profiles of $\dual L$ and of $\dual{\widetilde L}$ at $e^\eps$ are $C(e^\eps)$ and $\widetilde C(e^\eps)$. For a knot $s\in S$, the function $x\mapsto(x-s)_+$ is affine between every two adjacent knots, and the CtD projection preserves the mean of every value. Thus $\widetilde C(s)=C(s)$. Between two consecutive knots $u<v$, every value of $\widetilde\lr$ is either at most $u$ or at least $v$. Thus $\widetilde C$ is affine on $[u,v]$. In addition, $\widetilde C(0)=C(0)=1$, and $\widetilde C(s)=1-\expect{}{\lr}$ for $s\ge s_{\max}$ since $\widetilde\lr\le s_{\max}$. This proves the first part.

For the second part, $(1-e^\eps x)_+=e^\eps\bigl((x-e^{-\eps})_+-x+e^{-\eps}\bigr)$. Thus by~\eqref{eq:profiles} the privacy profile of $L$ at $e^\eps$ is $1-e^\eps+e^\eps C(e^{-\eps})$, and the same holds for $\widetilde L$ and $\widetilde C$. Thus the two privacy profiles are equal at every point $e^\eps=1/s$ with $s\in S\setminus\{0\}$. If $\widetilde C(s)=A+Bs$ for $s=e^{-\eps}$ between two consecutive knots, then the privacy profile of $\widetilde L$ is $(1+B)+(A-1)e^\eps$, which is affine in $e^\eps$. For $e^\eps\le1/s_{\max}$ it is $1-\expect{}{\lr}e^\eps$. If $S$ has a smallest positive knot $s_{\min}$, then $\widetilde C$ is affine on $[0,s_{\min}]$ with $\widetilde C(0)=1$, that is, $A=1$ there, and the privacy profile is constant for $e^\eps\ge1/s_{\min}$.

Since $\widetilde L$ is supported on $-\ln(S\setminus\{0\})\cup\{\infty\}$ and, by the definition of the dual, $\dual{\widetilde L}$ is supported on $\ln(S\setminus\{0\})\cup\{\infty\}$, the two parts of the claim identify them with the pessimistic CtD approximations defined above (see also \citep[Remark~C.2]{FS26pld}).
\end{proof}
In particular, the CtD approximations of $L$ and of $\dual L$ obtained in this way are duals of each other. FS26 require this property when they use CtD to discretize the input of the remove direction \citep[Remark~C.2]{FS26pld}.

\subsection{Couplings and the Proof of Theorem~\ref{thm:ctd-discretized}}\label{apd:ctd-thm}
\begin{lemma}[Coupling along the tree]\label{lem:tree-coupling}
Consider an unfolded tree of depth $d$ whose leaves are independent nonnegative random variables, in which every leaf applies the CtD projection onto $\grid$, as an independent random map, to its value, and every internal node applies it to the sum of the values of its children. Let $\mathcal L$ denote the values of the leaves, let $Y$ be their sum and let $\widehat Y$ be the value at the root. Then
\begin{equation}\label{eq:tree-coupling}
 \expect{}{\widehat Y\mid\mathcal L}=Y,
 \qquad
 \Bigl(1+\frac1m\Bigr)^{-(d+1)}Y\le\widehat Y\le\Bigl(1+\frac1m\Bigr)^{d+1}Y,
\end{equation}
and $\widehat Y=0$ if and only if $Y=0$.
\end{lemma}
\begin{proof}
We prove the claims for every node $n$ by induction on its height $h(n)$, where $Y_n$ and $\widehat Y_n$ denote the exact and the computed sums at $n$. For a leaf, $\widehat Y_n$ is the projection of $Y_n$, and by~\eqref{eq:projection-coupling}, $\expect{}{\widehat Y_n\mid\mathcal L}=Y_n$ and $\widehat Y_n$ is within a factor of $1+1/m$ of $Y_n$. For an internal node with children $c_1$ and $c_2$, the input of the projection is $y_n=\widehat Y_{c_1}+\widehat Y_{c_2}$. By the induction hypothesis, $\expect{}{y_n\mid\mathcal L}=Y_{c_1}+Y_{c_2}=Y_n$ and $y_n$ is within a factor of $(1+1/m)^{h(n)}$ of $Y_n$. By~\eqref{eq:projection-coupling}, $\expect{}{\widehat Y_n\mid y_n,\mathcal L}=y_n$ and $\widehat Y_n$ is within a factor of $1+1/m$ of $y_n$, and thus within a factor of $(1+1/m)^{h(n)+1}$ of $Y_n$. This gives~\eqref{eq:tree-coupling} at the root. Finally, the projection maps zero to itself and positive values to positive values (Claim~\ref{clm:random-map}).
\end{proof}

Next we show how to convert a coupling of the likelihood ratios under $Q$ into couplings of the PLDs in both directions. Note that this conversion does not introduce a multiplicative factor in the error in the remove direction.

\begin{lemma}[Two directions]\label{lem:two-directions}
Let $Y$ and $\widehat Y$ be bounded nonnegative random variables on a common probability space such that $\expect{}{Y}\le1$ and
\[
 \expect{}{\widehat Y\mid Y}=Y,
 \qquad e^{-\gamma}Y\le\widehat Y\le e^\gamma Y.
\]
Then $-\ln(Y)$ and $-\ln(\widehat Y)$ are PLD realizations, and for $(A,\widehat A)=(-\ln(Y),-\ln(\widehat Y))$ and for $(A,\widehat A)=(\dual{-\ln(Y)},\dual{-\ln(\widehat Y)})$ there is a coupling of $A$ and $\widehat A$ under which $A=\infty$ if and only if $\widehat A=\infty$, and $|A-\widehat A|\le\gamma$ otherwise. In particular, $A\domSt_{(\gamma,0)}\widehat A$ and $\widehat A\domSt_{(\gamma,0)}A$.
\end{lemma}
\begin{proof}
Since $\expect{}{\widehat Y}=\expect{}{Y}\le1$, both are PLD realizations. Under a coupling as in the statement, $\widehat A\le A+\gamma$ and $A\le\widehat A+\gamma$, which gives the last statement. We now construct the couplings. Let $\pi$ be the joint distribution of $(Y,\widehat Y)$. For $-\ln(Y)$ and $-\ln(\widehat Y)$, the coupling is the joint distribution of $-\ln(Y)$ and $-\ln(\widehat Y)$ under $\pi$. Indeed, $Y=0$ if and only if $\widehat Y=0$, and the finite values differ by at most $\gamma$. For the duals, we weight $\pi(dy,d\hat y)$ by $\hat y$. Since $\expect{}{\widehat Y\mid Y}=Y$, the marginal of the resulting measure over $y$ is the distribution of $Y$ weighted by $y$, and its marginal over $\hat y$ is the distribution of $\widehat Y$ weighted by $\hat y$. This measure has mass $\expect{}{Y}$, and under it both coordinates are positive. Taking the logarithm of both coordinates and adding mass $1-\expect{}{Y}$ at $(\infty,\infty)$, we obtain a coupling of the two duals (by the definition of the dual) in which the finite values differ by at most $\gamma$.
\end{proof}

\begin{claim}\label{clm:bands}
In the setting of Theorem~\ref{thm:ctd-discretized}, the positive finite part of the distribution computed at a node of height $h$ of the unfolded tree of repeated squaring is supported on the bands $\indLo,\ldots,\indHi+h+1$. In particular, every distribution in the computation is supported on at most $J+\lceil\log_2(t)\rceil+1$ bands.
\end{claim}
\begin{proof}
We use induction on $h$. For a leaf ($h=0$) this holds since the positive values of $e^{-L'}$ are knots of $\grid$ in $[2^{\indLo},2^{\indHi+1}]$ (Section~\ref{sec:ctd-alloc}), and $2^{\indHi+1}$ is the smallest knot of band $\indHi+1$. The children of a node of height $h\ge1$ have heights at most $h-1$, and thus by the induction hypothesis their positive finite parts are supported on the bands $\indLo,\ldots,\indHi+h$. By Theorem~\ref{thm:merge}, the positive finite part of their merge is supported on the bands $\indLo,\ldots,\indHi+h+1$. The last statement holds since $h\le\lceil\log_2(t)\rceil$ (Claim~\ref{clm:depth}) and $J=\indHi-\indLo+1$.
\end{proof}

\begin{proof}[Proof of Theorem~\ref{thm:ctd-discretized}]
\emph{Setup.} By line~2, the positive values of $e^{-L}$ lie in $[2^{\indLo},2^{\indHi+1}]$, and thus $e^{-L}$ is a bounded nonnegative random variable. Consider the unfolded tree of repeated squaring for the sum of $t$ copies, with independent copies of $e^{-L}$ at its leaves, in which every leaf projects its value onto $\grid$ (line~3) and every projection is applied as an independent random map. The projected leaves are independent copies of $e^{-L'}$, whose positive values are knots of $\grid$ in $[2^{\indLo},2^{\indHi+1}]$. The depth of the tree is $d\dfn\lceil\log_2(t)\rceil$ (Claim~\ref{clm:depth}), and the value $\sumHat$ at its root has the distribution $f_{\sumHat}$ computed in line~4 of Algorithm~\ref{alg:ctd-alloc} (Claim~\ref{clm:tree-dist}). Let $\lrAvg t\dfn\frac1t\sum_{i=1}^te^{-L^{(i)}}$ be the average of the leaves and let $\lrHat\dfn\sumHat/t$. By line~5, $\add L_t=-\ln(\lrHat)$ and $\rem L_t=\dual{-\ln(\lrHat)}$, and by~\eqref{eq:psi} and~\eqref{eq:alloc-dual}, $\psiAdd t(L)=-\ln(\lrAvg t)$ and $\psiRem t(\dual L)=\dual{-\ln(\lrAvg t)}$.

\emph{Validity.} Let $\gamma=(d+1)\ln(1+1/m)$. By Lemma~\ref{lem:tree-coupling}, $\expect{}{\sumHat\mid e^{-L^{(1)}},\ldots,e^{-L^{(t)}}}=t\lrAvg t$ and $e^{-\gamma}t\lrAvg t\le\sumHat\le e^{\gamma}t\lrAvg t$. Dividing by $t$, and since $\lrAvg t$ is a function of the leaves, the tower property gives $\expect{}{\lrHat\mid\lrAvg t}=\lrAvg t$ and $e^{-\gamma}\lrAvg t\le\lrHat\le e^{\gamma}\lrAvg t$. In particular, $\lrAvg t\cx\lrHat$ by Jensen's inequality. Since $e^{-L}$ and $\lrHat$ are bounded and nonnegative, Lemma~\ref{lem:validity} shows that $\add L_t$ is a PLD realization that dominates $\psiAdd t(L)$, and that $\rem L_t$ dominates $\psiRem t(\dual L)$. Since the dual of a PLD realization is a PLD realization (Claim~\ref{clm:pld-real}), $\rem L_t$ is also a PLD realization.

\emph{Tightness.} The variables $\lrAvg t$ and $\lrHat$ are bounded and nonnegative, and $\expect{}{\lrAvg t}=\expect{}{e^{-L}}\le1$. Thus Lemma~\ref{lem:two-directions} with $(Y,\widehat Y)=(\lrAvg t,\lrHat)$ gives $\add L_t\domSt_{(\gamma,0)}\psiAdd t(L)$ and $\rem L_t\domSt_{(\gamma,0)}\psiRem t(\dual L)$. By Claim~\ref{clm:stoch-hs}, $\HockeyStickFunc{e^\eps}{\add L_t}\le\HockeyStickFunc{e^{\eps-\gamma}}{\psiAdd t(L)}$ and $\HockeyStickFunc{e^\eps}{\rem L_t}\le\HockeyStickFunc{e^{\eps-\gamma}}{\psiRem t(\dual L)}$ for every $\eps\in\reals$. Together with validity, these are the two inequalities of Definition~\ref{def:tight-dom} with the parameters $(\gamma,0)$.

\emph{$\chi^2$ error and bias.} By the setup, Lemma~\ref{lem:chi2-tree} applies to this tree, and its exact and computed sums at the root are $Y=t\lrAvg t$ and $\widehat Y=\sumHat$. Thus its two pairs $(A,\widehat A)$ are $(\psiAdd t(L),\add L_t)$ and $(\psiRem t(\dual L),\rem L_t)$, and for each of them with $\chi^2(A)<\infty$ the $\chi^2$ error is at most $((1+\tau)^{d+1}-1)(1+\chi^2(A))$ for $\tau\dfn1/(4m(m+1))$. For $m\ge d+1$ we have $(d+1)\tau\le1$, and thus
\[
 (1+\tau)^{d+1}-1\le e^{(d+1)\tau}-1\le(d+1)\tau+(d+1)^2\tau^2\le\frac{d+1}{4m^2},
\]
where the last inequality holds since $d+1\le4(m+1)$. This gives the bound on the $\chi^2$ error, and Appendix~\ref{apd:bias} gives the bound on the bias.

\emph{Computational complexity.} The positive finite part of every distribution in the computation is supported on at most $J+d+1$ bands (Claim~\ref{clm:bands}). Repeated squaring performs at most $2d$ merges (Claim~\ref{clm:depth}), and by Theorem~\ref{thm:merge}, each of them uses $O((J+d)m\log(m))$ operations and $O((J+d)m)$ space. Besides the input, repeated squaring keeps at most two intermediate distributions, each supported on $O((J+d)m)$ points. Line~5 computes both outputs from $f_{\sumHat}$ in time linear in the size of its support, that is, $O((J+d)m)$. Since $d=O(\log(t))$, the running time is $O((J+\log(t))\,m\log(m)\log(t))$ and the space is $O((J+\log(t))\,m)$, apart from the computation of $L'$ in lines~2 and~3, which depends on the representation of the input.
\end{proof}

\subsection{\texorpdfstring{$\chi^2$}{Chi-Squared} Error of a CtD Projection}\label{apd:chi2-projection}
We first bound the $\chi^2$ error of rounding, which Section~\ref{sec:chi2} uses for the comparison with direct convolution.

\begin{claim}\label{clm:chi2-stoch}
Let $L$ and $\upBnd$ be PLD realizations with $\chi^2(L)<\infty$.
\begin{enumerate}
\item If $L\domSt\upBnd\domSt_{(\alpha,0)}L$ for some $\alpha\ge0$, then the $\chi^2$ error of $\upBnd$ relative to $L$ is between $0$ and $(e^\alpha-1)(1+\chi^2(L))$.
\item If $\upBnd$ is obtained by rounding $L$ up to the grid $\lossPar(\mathbb Z+\theta)$, where $\lossPar>0$ and $\theta$ is uniform on $[0,1)$ and independent of $L$, then the $\chi^2$ error of $\upBnd$ relative to $L$ is $((e^{\lossPar}-1)/\lossPar-1)(1+\chi^2(L))\ge(\lossPar/2)(1+\chi^2(L))$.
\end{enumerate}
\end{claim}
\begin{proof}
For the first part, $\overline F_L(x)\le\overline F_\upBnd(x)\le\overline F_L(x-\alpha)$ for all $x$. Since the exponential function is increasing, $\expect{}{e^L}\le\expect{}{e^\upBnd}\le\expect{}{e^{L+\alpha}}=e^\alpha\expect{}{e^L}$.

For the second part, $L$ is finite since $\chi^2(L)<\infty$. Let $Z\dfn L/\lossPar-\theta$, so that $\upBnd-L=\lossPar(\lceil Z\rceil-Z)$. Conditioned on $L$, the fractional part of $Z$ is uniform on $[0,1)$, and thus $\lceil Z\rceil-Z$ is uniform on $[0,1)$ up to an event of probability zero. Therefore $\expect{}{e^\upBnd\mid L}=e^L\int_0^1e^{\lossPar v}\,dv=e^L(e^{\lossPar}-1)/\lossPar$, and $(e^{\lossPar}-1)/\lossPar-1=\sum_{k\ge1}\lossPar^k/(k+1)!\ge\lossPar/2$.
\end{proof}
In contrast, the CtD projection changes the $\chi^2$ divergence only to the second order.

\begin{lemma}[$\chi^2$ error of a CtD projection]\label{lem:chi2-projection}
Let $\lossPar>0$ and let $S\subseteq[0,\infty)$ be a set of knots that contains $0$ and in which every two consecutive positive knots $u<v$ satisfy $v\le e^{\lossPar}u$. Let $\lr$ be a nonnegative random variable whose positive values lie between positive knots of $S$, and let $\widetilde\lr$ be its CtD projection onto $S$ (as a random map). Then
\[
 \expect{}{\lr^2}\le\expect{}{\widetilde\lr^2}\le\cosh^2(\lossPar/2)\,\expect{}{\lr^2}
\]
and
\[
 \expect{}{\lr^{-1}\ind_{\{\lr>0\}}}\le\expect{}{\widetilde\lr^{-1}\ind_{\{\widetilde\lr>0\}}}\le\cosh^2(\lossPar/2)\,\expect{}{\lr^{-1}\ind_{\{\lr>0\}}}.
\]
In particular, let $L$ be a PLD realization, and let $\widetilde L$ be the PLD realization for which $f_{e^{-\widetilde L}}=\proj_S(f_{e^{-L}})$. Then for $(A,\widetilde A)=(L,\widetilde L)$ and for $(A,\widetilde A)=(\dual L,\dual{\widetilde L})$ with $\chi^2(A)<\infty$, the $\chi^2$ error of $\widetilde A$ relative to $A$ is between $0$ and $\sinh^2(\lossPar/2)(1+\chi^2(A))$. The factor $\cosh^2(\lossPar/2)$ cannot be improved. If $v=e^{\lossPar}u$ are adjacent knots of $S$ and $\lr\equiv(u+v)/2$, then the second upper bound holds with equality.
\end{lemma}
\begin{proof}
The projection fixes $0$ and the knots, and by \refTag{P1} it maps positive values to positive values. Thus it suffices to consider a value $x$ of $\lr$ between adjacent positive knots $u<x<v$, where $v\le e^{\lossPar}u$. The projection maps $x$ to $u$ and to $v$ with the probabilities $(v-x)/(v-u)$ and $(x-u)/(v-u)$, and thus with mean $x$ and variance $(x-u)(v-x)$. Direct calculation gives
\[
 \frac{\expect{}{\widetilde\lr^2\mid\lr=x}}{x^2}=1+\frac{(x-u)(v-x)}{x^2}
 \qquad\text{and}\qquad
 \frac{\expect{}{\widetilde\lr^{-1}\mid\lr=x}}{x^{-1}}=1+\frac{(x-u)(v-x)}{uv}.
\]
Since $(x-u)(v-x)\le(v-u)^2/4$, the second fraction after $1$ is at most $(v-u)^2/(4uv)$, and so is the first, since $(v-u)^2x^2-4uv(x-u)(v-x)=((u+v)x-2uv)^2\ge0$. In addition, $1+(v-u)^2/(4uv)=\cosh^2(\ln(v/u)/2)\le\cosh^2(\lossPar/2)$. Integrating over the values of $\lr$ gives the two chains of inequalities. For $x=(u+v)/2$ and $v=e^{\lossPar}u$, the second fraction is equal to $(v-u)^2/(4uv)$, which gives the equality.

For the second statement, let $\lr\dfn e^{-L}$. Since $\chi^2(A)<\infty$, $A$ has no mass at infinity. By \refTag{P1} and \refTag{P5}, $\prob{}{\widetilde\lr=0}=\prob{}{\lr=0}$ and $\expect{}{\widetilde\lr}=\expect{}{\lr}$, and thus $\widetilde A$ has no mass at infinity either. Therefore the definition of the dual gives $\expect{}{e^{L}}=\expect{}{\lr^{-1}}$ for $A=L$ and $\expect{}{e^{\dual L}}=\expect{}{\lr^2}$ for $A=\dual L$, and similarly for $\widetilde A$. Thus the two chains of inequalities give the claim, since $\cosh^2(\lossPar/2)-1=\sinh^2(\lossPar/2)$.
\end{proof}
For the grid $\grid$ we can take $\lossPar=\ln(1+1/m)$, for which $\sinh^2(\lossPar/2)=1/(4m(m+1))$. Thus for a single projection onto $\grid$, the $\chi^2$ error of $\widetilde A$ relative to $A$ is at most $(1+\chi^2(A))/(4m(m+1))$.

\subsection{\texorpdfstring{Proof of the $\chi^2$ Error Bound}{Proof of the Chi-Squared Error Bound}}\label{apd:chi2-proof}
\begin{lemma}[$\chi^2$ error along the tree]\label{lem:chi2-tree}
Consider the unfolded tree of Lemma~\ref{lem:tree-coupling} of depth $d$ with $t$ leaves, which are independent copies of $e^{-L}$ for a PLD realization $L$ such that $e^{-L}$ takes values in $\{0\}\cup[2^{\indLo},2^{\indHi+1}]$ for integers $\indLo\le\indHi$. Let $Y$ and $\widehat Y$ denote the exact and the computed sums at the root, and let $\tau\dfn1/(4m(m+1))$. Then
\[
 \expect{}{Y^2}\le\expect{}{\widehat Y^2}\le(1+\tau)^{d+1}\,\expect{}{Y^2}
\]
and
\[
 \expect{}{Y^{-1}\ind_{\{Y>0\}}}\le\expect{}{\widehat Y^{-1}\ind_{\{\widehat Y>0\}}}\le(1+\tau)^{d+1}\,\expect{}{Y^{-1}\ind_{\{Y>0\}}}.
\]
In particular, for $(A,\widehat A)=(-\ln(Y/t),-\ln(\widehat Y/t))$ and for $(A,\widehat A)=(\dual{-\ln(Y/t)},\dual{-\ln(\widehat Y/t)})$, if $\chi^2(A)<\infty$, then the $\chi^2$ error of $\widehat A$ relative to $A$ is between $0$ and $((1+\tau)^{d+1}-1)(1+\chi^2(A))$.
\end{lemma}
We prove Lemma~\ref{lem:chi2-tree} by induction on the subtrees of the unfolded tree. All the random variables below are bounded, and on the event that they are positive they are bounded away from zero. Thus all the expectations are finite. Let $\varphi(w)=w^2$ or $\varphi(w)=1/w$, which are convex on $(0,\infty)$ and satisfy $\varphi(ab)=\varphi(a)\varphi(b)$. For a node $n$, let $Y_n$ and $\widehat Y_n$ denote the exact and the computed sums at $n$, and let $\mathcal L$ denote the values of the leaves, as in Lemma~\ref{lem:tree-coupling}.

\begin{claim}[Moments along the subtrees]\label{clm:subtree-chi2}
For every node $n$ of height $h$, on the event $Y_n>0$,
\[
 1\le\expect{}{\varphi(\widehat Y_n/Y_n)\mid\mathcal L}\le(1+\tau)^{h+1}.
\]
\end{claim}
\begin{proof}
The lower bound follows from Jensen's inequality, since $\expect{}{\widehat Y_n\mid\mathcal L}=Y_n$ by Lemma~\ref{lem:tree-coupling} applied to the subtree of $n$. By Lemma~\ref{lem:chi2-projection} applied to a constant $y>0$, with $\lossPar=\ln(1+1/m)$ by \refTag{G1}, the projection $\tilde y$ of $y$ satisfies $\expect{}{\varphi(\tilde y/y)}\le\cosh^2(\lossPar/2)=1+\tau$. At a leaf, this is the upper bound for $h=0$. At an internal node $n$ of height $h$, conditioned on $\mathcal L$, the sums $y_i$ at its children $n_i$ are fixed, and the computed value at a child with $y_i=0$ is zero (Lemma~\ref{lem:tree-coupling}). Thus the input $s$ of the projection at $n$ satisfies
\[
 \frac s{Y_n}=\sum_{i:\,y_i>0}\frac{y_i}{Y_n}\cdot\frac{\widehat Y_{n_i}}{y_i},
\]
which is a convex combination. Since $\varphi$ is convex and the heights of the children are at most $h-1$, the induction hypothesis gives $\expect{}{\varphi(s/Y_n)\mid\mathcal L}\le(1+\tau)^h$. Conditioned also on $s$, the projection at $n$ gives $\expect{}{\varphi(\widehat Y_n/s)\mid s,\mathcal L}\le1+\tau$. Since $\widehat Y_n/Y_n=(\widehat Y_n/s)(s/Y_n)$ and $\varphi$ is multiplicative, the tower property gives the upper bound.
\end{proof}

\begin{proof}[Proof of Lemma~\ref{lem:chi2-tree}]
The root has height at most $d$. Multiplying Claim~\ref{clm:subtree-chi2} at the root by $Y^2$, or by $Y^{-1}$ on the event $Y>0$, and taking expectations gives the two chains of inequalities, where the reciprocal moments are restricted to the event $Y>0$, which is the event $\widehat Y>0$. For the second statement, by the definition of the dual, the masses of $-\ln(Y/t)$ and of its dual at infinity are $\prob{}{Y=0}$ and $1-\expect{}{Y}/t$, and the expectations of $e^{-\ln(Y/t)}$ and of the exponential of its dual over their finite parts are $t\,\expect{}{Y^{-1}\ind_{\{Y>0\}}}$ and $\expect{}{Y^2}/t^2$, and similarly for $\widehat Y$. By Lemma~\ref{lem:tree-coupling}, $\prob{}{\widehat Y=0}=\prob{}{Y=0}$ and $\expect{}{\widehat Y}=\expect{}{Y}$, and thus $A$ and $\widehat A$ have the same mass at infinity. If $\chi^2(A)<\infty$, this mass is zero, and the two chains of inequalities give the bounds on the $\chi^2$ errors.
\end{proof}
Lemma~\ref{lem:chi2-tree} gives the bound on the $\chi^2$ error in the third part of Theorem~\ref{thm:ctd-discretized}, since $-\ln(Y/t)=\psiAdd t(L)$ and $\dual{-\ln(Y/t)}=\psiRem t(\dual L)$ by~\eqref{eq:alloc-dual}.

\subsection{Proof of the Bias Bound}\label{apd:bias}
\begin{proof}[Proof of the bound on the bias in Theorem~\ref{thm:ctd-discretized}]
We use the unfolded tree of the proof of Theorem~\ref{thm:ctd-discretized}, whose depth is $d\dfn\lceil\log_2(t)\rceil$, and the notation $Y_n$, $\widehat Y_n$ and $\mathcal L$ of Appendix~\ref{apd:chi2-proof}, with $Y$ and $\widehat Y$ the sums at the root. Let $\lossPar\dfn\ln(1+1/m)$. By line~5 and~\eqref{eq:alloc-dual}, the allocation PLD and the output are $A=-\ln(Y/t)$ and $\widehat A=-\ln(\widehat Y/t)$ in the add direction, and their duals in the remove direction. As in the proof of Lemma~\ref{lem:chi2-tree}, $A$ and $\widehat A$ have the same mass at infinity.

We first bound the bias of a single projection. For the projection $\tilde y$ of a fixed value $y>0$, let $Z\dfn\tilde y/y$. By \refTag{P1}, \refTag{P5} and \refTag{G1}, $-\ln(\tilde y)$ is the CtD discretization of the loss $-\ln(y)$ on a grid with spacing at most $\lossPar$ \citep[Definition~6.1]{FS26clt}, which changes this loss by $-\ln(Z)$. In the dual, the loss $\ln(y)$ changes by $\ln(Z)$, with the distribution of $Z$ weighted by $Z$. Thus the bound on the bias of a CtD discretization \citep[Lemma~6.2 and its proof]{FS26clt}, applied in both directions, gives
\[
 0\le\expect{}{-\ln(Z)}\le\frac{\lossPar^2}8
 \qquad\text{and}\qquad
 0\le\expect{}{Z\ln(Z)}\le\frac{\lossPar^2}8 .
\]

Let $\varphi(w)=-\ln(w)$ or $\varphi(w)=w\ln(w)$, which are convex on $(0,\infty)$ and satisfy $\varphi(1)=0$. We show that for every node $n$ of height $h$, on the event $Y_n>0$,
\begin{equation}\label{eq:bias-conditional}
 0\le\expect{}{\varphi(\widehat Y_n/Y_n)\mid\mathcal L}\le\frac{(h+1)\lossPar^2}{8}.
\end{equation}
The lower bound follows from Jensen's inequality, since $\expect{}{\widehat Y_n\mid\mathcal L}=Y_n$ by Lemma~\ref{lem:tree-coupling} applied to the subtree of $n$. We prove the upper bound by induction on $h$, as in the proof of Claim~\ref{clm:subtree-chi2}. At a leaf, it is the bound for a single projection. At an internal node $n$, conditioned on $\mathcal L$, the ratio $V\dfn s/Y_n$ for the input $s$ of the projection at $n$ is a convex combination of the ratios $\widehat Y_{n_i}/Y_{n_i}$ at the children $n_i$ of $n$ with $Y_{n_i}>0$. Since $\varphi$ is convex and the heights of the children are at most $h-1$, the induction hypothesis gives $\expect{}{\varphi(V)\mid\mathcal L}\le h\lossPar^2/8$. Conditioned also on $V$, the ratio $\widehat Y_n/s$ is the ratio $Z$ for $y=s$, and $\widehat Y_n/Y_n=VZ$. Therefore
\[
 \expect{}{-\ln(VZ)\mid V,\mathcal L}\le-\ln(V)+\frac{\lossPar^2}8,
 \qquad
 \expect{}{VZ\ln(VZ)\mid V,\mathcal L}\le V\ln(V)+\frac{V\lossPar^2}8 .
\]
Since $\expect{}{V\mid\mathcal L}=1$, taking expectations conditioned on $\mathcal L$ gives the upper bound.

The root has height at most $d$. In the add direction, both losses are finite exactly when $Y>0$, and then $\widehat A-A=-\ln(\widehat Y/Y)$. Thus the bias is the expectation of $\expect{}{\varphi(\widehat Y/Y)\mid\mathcal L}\ind_{\{Y>0\}}$ for $\varphi(w)=-\ln(w)$. In the remove direction, the finite parts of $A$ and $\widehat A$ are distributed as $\ln(Y/t)$ and $\ln(\widehat Y/t)$, weighted by $Y/t$ and $\widehat Y/t$, respectively (Section~\ref{sec:prelims}). Since $\expect{}{\widehat Y\mid\mathcal L}=Y$, the bias is then the expectation of $(Y/t)\,\expect{}{\varphi(\widehat Y/Y)\mid\mathcal L}\ind_{\{Y>0\}}$ for $\varphi(w)=w\ln(w)$. Therefore~\eqref{eq:bias-conditional} and $\expect{}{Y/t}\le1$ give the bounds on the bias.
\end{proof}

\subsection{Guarantees for a Given Accuracy}\label{apd:input}
\begin{proof}[Proof of Corollary~\ref{cor:ctd-original}]
The algorithm returns the add output of Algorithm~\ref{alg:ctd-alloc} on $\add L$ and the remove output of Algorithm~\ref{alg:ctd-alloc} on $\dual{\rem L}$, since $\psiRem t(\dual{\dual{\rem L}})=\psiRem t(\rem L)$. (If $\rem L$ and $\add L$ are duals of each other, then a single call gives both outputs.) Consider one of these calls, and denote its input by $L$ and its outputs by $\add L_t$ and $\rem L_t$. The bands chosen in line~2 satisfy $\indHi-\indLo\le(\ellHi-\ellLo)/\ln(2)+1\le\wid/\ln(2)+1$, since the finite values of $-\dual{\rem L}$ are those of $\rem L$. Let $\lossPar=\ln(1+1/m)$, so that $(\lceil\log_2(t)\rceil+1)\lossPar\le(\lceil\log_2(t)\rceil+1)/m\le\alpha$. Thus by the first two parts of Theorem~\ref{thm:ctd-discretized}, $\add L_t$ and $\rem L_t$ $(\alpha,0)$-tightly dominate $\psiAdd t(L)$ and $\psiRem t(\dual L)$.

\emph{$\chi^2$ error and bias.} Since $m\ge(\lceil\log_2(t)\rceil+1)/\alpha\ge\lceil\log_2(t)\rceil+1$, the third part of Theorem~\ref{thm:ctd-discretized} bounds the $\chi^2$ error of each output relative to the corresponding allocation PLD $\Psi$ of $L$ with $\chi^2(\Psi)<\infty$ by $(\lceil\log_2(t)\rceil+1)(1+\chi^2(\Psi))/(4m^2)\le\alpha^2(1+\chi^2(\Psi))/(4(\lceil\log_2(t)\rceil+1))$, and the bias of each output by $(\lceil\log_2(t)\rceil+1)\lossPar^2/8\le\alpha^2/(8(\lceil\log_2(t)\rceil+1))$.

\emph{Complexity.} Since $\indHi-\indLo\le\wid/\ln(2)+1$, the positive values of $e^{-L'}$ lie in the $J+1=O(1+\wid)$ bands $\indLo,\ldots,\indHi+1$. Computing $\dual{\rem L}$ and $L'$ takes $O(S+Jm)$ operations, since every atom contributes to at most two knots. Substituting $m=O(\log(t)/\alpha)$ into Theorem~\ref{thm:ctd-discretized} gives the bounds on the running time and the space.
\end{proof}

\printbibliography

\end{document}